\documentclass{article}
\pdfoutput=1

\PassOptionsToPackage{numbers, compress}{natbib}

\usepackage[final]{neurips_2025}

\usepackage[utf8]{inputenc} 
\usepackage[T1]{fontenc}    
\usepackage{hyperref}       
\usepackage{url}            
\usepackage{booktabs}       
\usepackage{amsfonts}       
\usepackage{nicefrac}       
\usepackage{microtype}      
\usepackage{xcolor}         
\usepackage{graphicx}

\usepackage{multirow}
\usepackage{graphicx}
\usepackage[table]{xcolor}
\usepackage{adjustbox}
\usepackage{amssymb}
\usepackage{amsmath}
\usepackage{stmaryrd}
\usepackage{dsfont}
\usepackage{subcaption}
\usepackage{pifont}
\usepackage{wrapfig}
\usepackage{adjustbox}
\usepackage{tabularx}
\usepackage{makecell}
\usepackage[dvipsnames]{xcolor}

\newcommand{\ie}{\textit{i}.\textit{e}., }
\newcommand{\eg}{\textit{e}.\textit{g}., }

\title{DeepVideo-R1: Video Reinforcement Fine-Tuning\\via Difficulty-aware Regressive GRPO}

\author{Jinyoung Park$^1$\thanks{Work was done at Korea University.} \quad Jeehye Na$^{1*}$ \quad Jinyoung Kim$^2$ \quad Hyunwoo J. Kim$^1$\thanks{corresponding author.}\\
  $^1$Korea Advanced Institute of Science and Technology, \quad $^2$Korea University\\
  {\tt \{jinyoung.park, jeehyena, hyunwoojkim\}@kaist.ac.kr} \\
  {\tt k012100@korea.ac.kr}\\
\textcolor{blue}{\url{https://github.com/mlvlab/DeepVideoR1}}}

\hypersetup{
  pdfauthor={Jinyoung Park, Jeehye Na, Jinyoung Kim, Hyunwoo J. Kim},
  pdftitle={DeepVideo-R1: Video Reinforcement Fine-Tuning via Difficulty-aware Regressive GRPO}
}

\begin{document}

\maketitle
\begin{abstract}
Recent works have demonstrated the effectiveness of reinforcement learning~(RL)-based post-training for enhancing the reasoning capabilities of large language models~(LLMs).
In particular, Group Relative Policy Optimization~(GRPO) has shown impressive success using a PPO-style reinforcement learning algorithm with group-normalized rewards.
However, the effectiveness of GRPO in Video Large Language Models~(VideoLLMs) remains underexplored.
In this paper, we explore GRPO and identify two issues that hinder effective learning: (1) reliance on safeguards, and (2) vanishing advantage.
To mitigate these challenges, we propose \textbf{DeepVideo-R1}, a video large language model trained with \textbf{Reg-GRPO}~(\textbf{Reg}ressive \textbf{GRPO}) and difficulty-aware data augmentation.
Reg-GRPO reformulates the GRPO loss function as a regression task that directly predicts the advantage in GRPO, eliminating the need for safeguards such as clipping and min operations.
This directly aligns the model with the advantages, providing guidance to prefer better outputs.
The difficulty-aware data augmentation strategy augments input prompts/videos to target solvable difficulty levels, enabling diverse reward signals.
Our experimental results show that our approach significantly improves video reasoning performance across multiple benchmarks.
\end{abstract}
\section{Introduction}
Large Language Models (LLMs)~\cite{radford2021learning,touvron2023llama,achiam2023gpt,team2023gemini} have demonstrated remarkable abilities in understanding, reasoning, and generating text across diverse domains.
Their success stems from next-token prediction over vast corpora, which enables the emergence of complex reasoning patterns and world knowledge.
Building on this progress, recent research has extended LLMs into the video domain, giving rise to Video Large Language Models (VideoLLMs)~\cite{wang2024qwen2,wang2024internvideo2,bai2025qwen2,zhu2025internvl3,zhang2025videollama}.
These models aim to unify video understanding and language generation, enabling capabilities such as temporal event reasoning, video question answering, and video-to-text summarization.

Despite their rapid evolution, current VideoLLMs still struggle with complex reasoning tasks, which require temporal, spatial, and semantic understanding over video sequences.
Since standard supervised fine-tuning fits instruction data rather than reasoning processes, it is limited in improving reasoning capabilities.
To address this, reinforcement learning (RL)-based post-training~\cite{zhang2024direct,ahn2025isr} has emerged as a compelling paradigm.
RL provides a mechanism to optimize models beyond likelihood objectives, aligning them with reward signals that encode human preference or task-specific success.
Recently, Group Relative Policy Optimization~(GRPO)~\cite{shao2024deepseekmath,guo2025deepseek} has shown promise by using group-based advantages and relative preference signals to enhance reasoning capabilities.

While GRPO has achieved strong results in text-based tasks, its application to VideoLLMs remains underexplored.
In this work, we investigate the application of GRPO to VideoLLMs and identify two key limitations that hinder effective training: (1) reliance on stabilizers such as minimum and clipping operations, which often suppress gradients and impede convergence, and (2) the vanishing advantage problem, where extremely easy or difficult samples yield zero advantages, thereby removing the training signal.

To overcome these limitations, we propose \textbf{DeepVideo-R1}, a video large language model trained with two key innovations: \textit{Regressive GRPO} (Reg-GRPO) and \textit{difficulty-aware data augmentation}.
Reg-GRPO reformulates the GRPO objective as a regression problem that directly predicts group-based advantage values.
This simple yet effective reformulation enables direct alignment between model outputs and the advantage values, eliminating the need for stabilizers while ensuring stable convergence.
We also introduce a difficulty-aware augmentation that dynamically adjusts the difficulty of video-text inputs.
For easy samples, we perturb the video content to inject uncertainty; for hard samples, we provide auxiliary reasoning cues.
This strategy diversifies the reward landscape, mitigating the vanishing advantage problem and promoting balanced learning across difficulty levels.

\begin{figure}[t!]
    \begin{center}
        {%
            \includegraphics[width=0.56\linewidth]{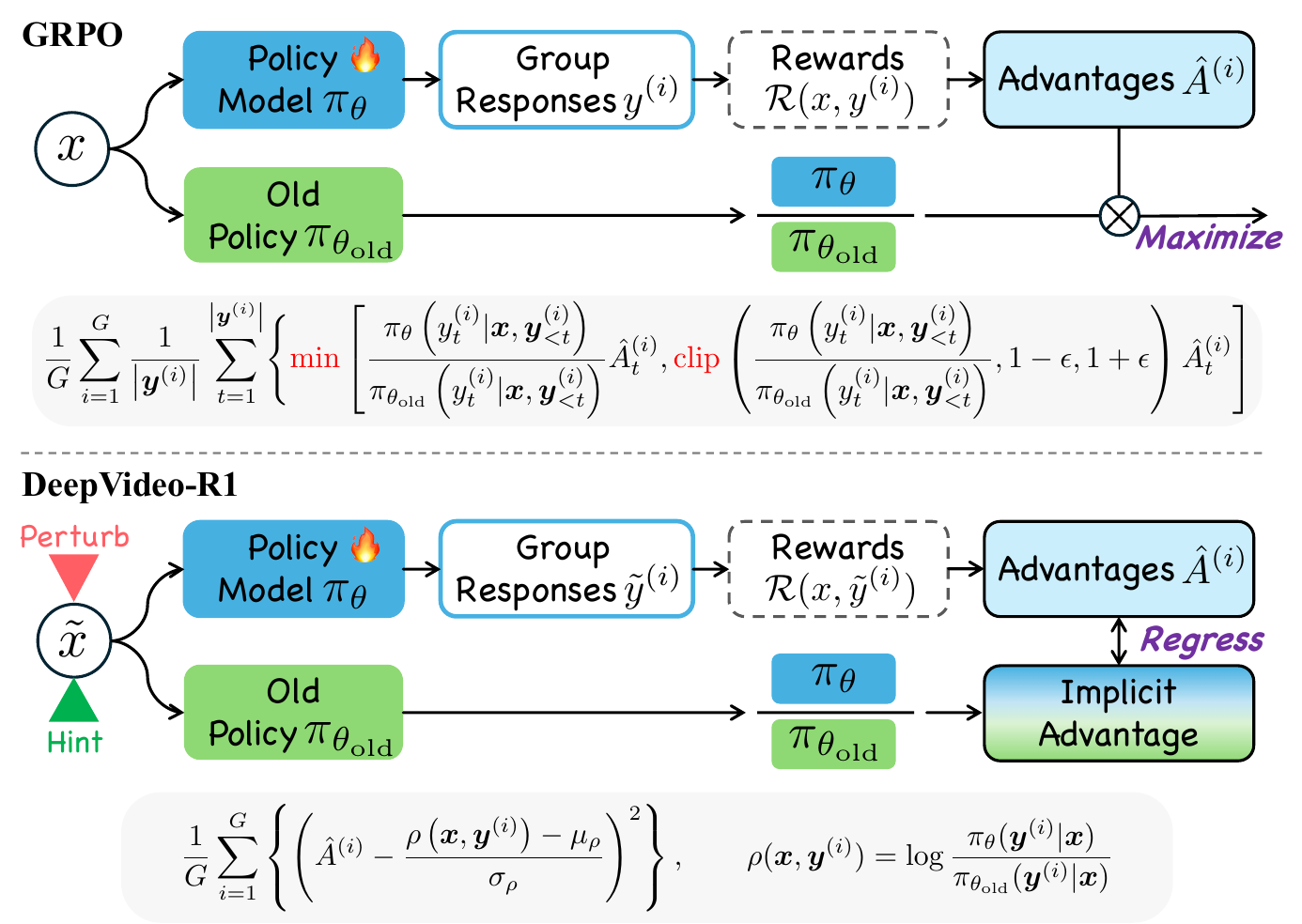}
            \label{fig:figure2_1}%
        }%
        \hfill%
        {%
            \includegraphics[width=0.42\linewidth]{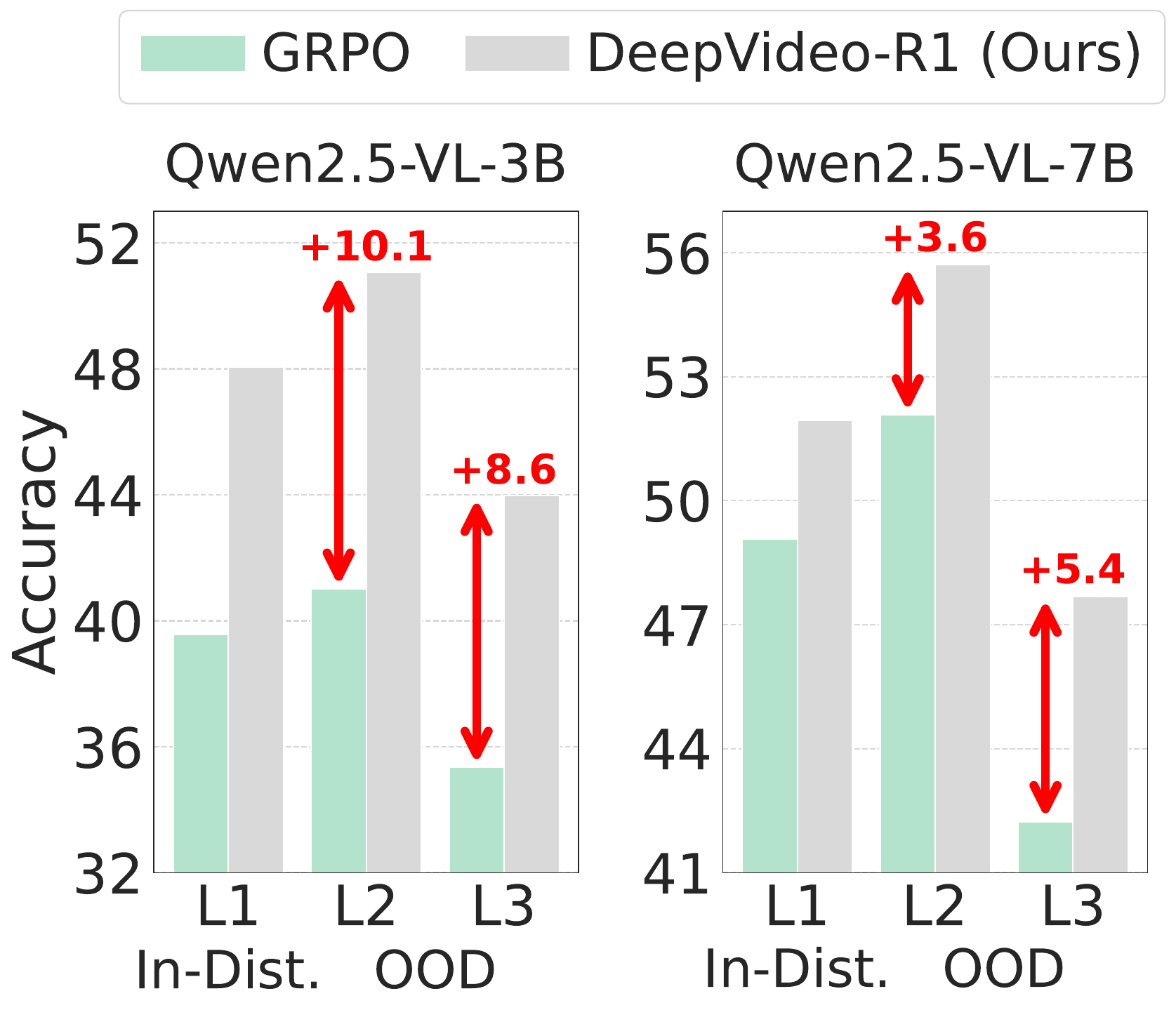}
            \label{fig:figure2_2}%
        }%
    \end{center}
\caption{\textbf{DeepVideo-R1 significantly improves the reasoning capabilities of VideoLLMs.} 
Our VideoLLM, DeepVideo-R1, is trained to explicitly predict the advantage $\hat{A}^{(i)}$ through Regressive GRPO loss.
Notably, model training becomes significantly effective and achieves a 10.1 performance improvement compared to GRPO.
}
\label{fig:Simple_Overview}
\end{figure}

Our experimental results demonstrate the effectiveness of DeepVideo-R1 on multiple challenging video reasoning benchmarks such as SEED-Bench-R1, LongVideoBench, and NExTGQA, achieving superior performance over recent video LLMs such as Qwen2.5-VL~\cite{bai2025qwen2}~(Figure~\ref{fig:Simple_Overview}).
Notably, our model achieves consistent improvements on both in-distribution and out-of-distribution tasks, indicating robust generalization capabilities.
These results underscore the benefits of combining a regression-based RL objective with data augmentation for training large-scale multimodal reasoning models.

Our main contributions are listed as:
\begin{itemize}
  \item We introduce \textbf{Reg-GRPO}, a novel optimization scheme that casts GRPO as a regression task over group-based advantage values, eliminating heuristic stabilizers such as clipping and min operations, and mitigating the vanishing gradient issue.
  \item We develop a \textbf{difficulty-aware augmentation} framework that modulates video-text inputs with adaptive difficulty scaling, video cue injection, and noise perturbation to generate richer and more effective reward signals.
  \item We propose \textbf{DeepVideo-R1}, a video large language model trained with two key innovations: Regressive GRPO~(Reg-GRPO) and difficulty-aware data augmentation. Our experimental results demonstrate that DeepVideo-R1 significantly improves the reasoning capabilities of VideoLLMs on complex video reasoning tasks.
\end{itemize}

\section{Related Work}
\paragraph{Video Large Language Models~(VideoLLMs).}
Large Language Models~(LLMs)~\cite{brown2020language,yang2024qwen2,grattafiori2024llama} have exhibited strong generalization and reasoning capabilities across a wide range of domains, including knowledge-intensive tasks~\cite{jin2025search,luo2025graph,park2023graph,park2024generative}, mathematical reasoning~\cite{shao2024deepseekmath,luo2023wizardmath}, and scientific domains~\cite{zheng2025large,bai2025intern,park2024llamo}.
Building on these capabilities, Video Large Language Models~(VideoLLMs) have extended LLM reasoning to dynamic video domains, achieving notable performance across tasks~\cite{wu2024longvideobench, fu2024video, li2024mvbench,shen2024longvu}, such as video question answering~\cite{bai2025qwen2, zhu2025internvl3, zhang2025videollama, li2023videochat,ko2023large,li2024llava} and video captioning~\cite{xu2023mplug, cheng2024videollama, zhang2024video}.
Despite their impressive performance, VideoLLMs remain limited on long video inputs and fine-grained video understanding tasks that require detailed spatiotemporal reasoning~\cite{gao2017tall, krishna2017dense, xiao2024can}.
Most existing methods primarily emphasize video perception or short-context understanding, often relying on static supervised fine-tuning objectives that fail to capture reasoning dynamics.
To address these challenges, we leverage a reinforcement learning-based fine-tuning approach to improve the reasoning and generalization capabilities of VideoLLMs.

\paragraph{RL-based fine-tuning.}
Multiple works~\cite{shao2024deepseekmath, guo2025deepseek,jaech2024openai, team2025kimi,lee2025vidchain,lee2025captioning} have significantly improved the reasoning capabilities of LLMs through reinforcement learning~(RL), including DPO~\cite{rafailov2023direct} and RLHF~\cite{ouyang2022training}.
Recently, variants of RL-based fine-tuning~\cite{zhu2024self,gao2024rebel} have explored direct reward-regression losses derived from RL objectives.
A key development in this direction is Group Relative Policy Optimization~(GRPO), an RL algorithm proposed in~\cite{shao2024deepseekmath} that computes group-wise normalized rewards to stabilize training and improve efficiency.
Motivated by GRPO, several approaches have demonstrated substantial improvements in the reasoning abilities of MLLMs across image~\cite{liu2025visual, yang2025r1, zhan2025vision, shen2025vlm, huang2025vision, huang2025boosting, liu2025noisyrollout, wang2025vl} and video tasks~\cite{feng2025video, li2025videochat, chen2025exploring, wang2025timezero, wu2025st, tan2025reason}.
While existing approaches~\cite{li2025videochat,wang2025timezero,chen2025exploring} have primarily focused on defining appropriate reward functions tailored to each visual task, some concurrent works~\cite{huang2025boosting, liu2025noisyrollout} instead study practical issues that arise during GRPO training, aiming to further enhance model reasoning.
In this work, we propose a learning algorithm that directly regresses advantages instead of increasing the likelihood of high-advantage responses.
Additionally, we design difficulty-aware data augmentation to provide diverse and dense learning signals.

\section{Methods}
In this section, we present a video large language model named \textcolor{black}{DeepVideo-R1}, which is trained with Regressive GRPO~(Reg-GRPO) and difficulty-aware data augmentation for effective video context reasoning.
We first introduce post-training methods for VideoLLMs, such as proximal policy optimization and group-relative policy optimization (GRPO), and discuss the limitations of GRPO: \emph{reliance on heuristic safeguards} and \emph{vanishing advantage}.
Then, we propose Reg-GRPO, which improves the RL-based GRPO by transforming the objective into a simpler yet more effective regression loss, eliminating heuristic safeguards such as the min and clipping operations.
Finally, we present a novel difficulty-aware data augmentation, which alleviates the vanishing advantage problem by modulating the difficulty of samples.
\subsection{RL-based Fine-Tuning}
\paragraph{Proximal Policy Optimization~(PPO)~\cite{wang2020truly}}
is one of the most widely used actor-critic RL algorithms for fine-tuning (video) large language models.
For example, RLHF~\cite{huang2024n+} applies the PPO algorithm.
Given the input sample $\boldsymbol{x}$, PPO optimizes the model $\pi_\theta$ with the following objective:
\begin{equation}
  \begin{split}
    &\mathcal{L}_{\text{PPO}}\left(\theta \right) = -\mathbb{E}_{\boldsymbol{x},\; \boldsymbol{y} \sim \pi_{\theta_{\text{old}}}\left(\cdot | \boldsymbol{x} \right)} \\
    &\frac{1}{\left\lvert \boldsymbol{y}\right\rvert}\sum_{t=1}^{\left\lvert\boldsymbol{y}\right\rvert}  \min \left[ \frac{\pi_\theta \left(y_t | \boldsymbol{x}, \boldsymbol{y}_{<t}  \right)}{\pi_{\theta_{\text{old}}} \left(y_t | \boldsymbol{x}, \boldsymbol{y}_{<t} \right)} {A}_t, \text{clip}\left(\frac{\pi_\theta \left(y_t | \boldsymbol{x}, \boldsymbol{y}_{<t}  \right)}{\pi_{\theta_{\text{old}}} \left(y_t | \boldsymbol{x}, \boldsymbol{y}_{<t} \right)},1-\epsilon, 1+\epsilon  \right)A_t \right] ,
  \end{split}
\end{equation}
where $\boldsymbol{y}$ is sampled from the policy model $\pi_\theta$, $\epsilon$ is a hyperparameter, $\pi_{\theta_\text{old}}$ is the old model and the advantage $A_t$ is calculated with generalized advantage estimation~(GAE)~\cite{schulman2015high} using rewards and a trained value function $V_\psi$.
Although the PPO algorithm aligns human preferences with the policy model outputs effectively, it requires substantial memory and computational resources since the value function is typically another model comparable in size to the policy model.

\paragraph{Group Relative Policy Optimization~(GRPO)~\cite{shao2024deepseekmath}}
addresses the problem of PPO~\cite{wang2020truly} by approximating the learnable value function with the average reward of multiple sampled outputs.
Concretely, given an input sample $\boldsymbol{x}$, the model samples multiple output sequences $\left\{\boldsymbol{y}^{(i)}\right\}_{i=1}^G$ from the old policy model $\pi_{\theta_\text{old}}$ and then trains the policy model $\pi_\theta$ with the following objective:
\begin{equation}
  \begin{split}
    &\mathcal{L}_{\text{GRPO}}\left(\theta \right) = \mathbb{E}_{\boldsymbol{x},\; \left\{\boldsymbol{y}^{(i)} \right\}_{i=1}^G \sim \pi_{\theta_{\text{old}}\left(\cdot | \boldsymbol{x} \right)}}\\
    &\frac{1}{\left\lvert \boldsymbol{y}^{(i)} \right\rvert} \sum_{t=1}^{\left\lvert \boldsymbol{y}^{(i)} \right\rvert} \Biggl\{ \min \left[  \frac{\pi_\theta \left(y_t^{(i)} | \boldsymbol{x}, \boldsymbol{y}_{<t}^{(i)}  \right)}{\pi_{\theta_{\text{old}}} \left(y_t^{(i)} | \boldsymbol{x}, \boldsymbol{y}_{<t}^{(i)} \right)} \hat{A}^{(i)}_t, \text{clip}\left( \frac{\pi_\theta \left(y_t^{(i)} | \boldsymbol{x}, \boldsymbol{y}_{<t}^{(i)}  \right)}{\pi_{\theta_{\text{old}}} \left(y_t^{(i)} | \boldsymbol{x}, \boldsymbol{y}_{<t}^{(i)} \right)},1-\epsilon, 1+\epsilon  \right)\hat{A}^{(i)}_t \right]
      \\
    &\quad \quad \quad \quad \quad - \beta \mathcal{D}_{\text{KL}}\left[\pi_\theta || \pi_{\text{ref}} \right]\Biggl\},
  \end{split}
\end{equation}
where $\beta$ corresponds to a hyperparameter and $\mathcal{D}_{\text{KL}}$ is the KL-divergence.
Here, $\hat{A}^{(i)}$ is advantage calculated based on the relative reward within the group, which is formulated as $\hat{A}^{(i)} = \frac{\mathcal{R}\left(\boldsymbol{x}, \boldsymbol{y}^{(i)} \right)-\mu_r}{\sigma_r}$
where $\mu_r, \sigma_r$ denotes the average and standard deviation values of a set of rewards in the group, respectively.
Although GRPO has shown its success, GRPO has two limitations that hinder the effective model optimization: \textit{reliance on heuristic constraints} and \textit{vanishing advantage} problems.

\paragraph{Reliance on safeguards.}
GRPO optimizes the model with safeguards implemented using the min and clipping functions to avoid extreme changes in the model.
However, the PPO-style clipping function induces \textbf{zero gradient} for samples where the value of ${\pi_\theta\left(\boldsymbol{y}|\boldsymbol{x} \right)}$ is too different from the value of ${\pi_{\theta_\text{old}}\left(\boldsymbol{y}|\boldsymbol{x} \right)}$.
It cannot guarantee that the model $\pi_{\theta}\left(\boldsymbol{y}|\boldsymbol{x}\right)$ stays close to $\pi_{\theta_{\text{ref}}}$ if it is already far from $\pi_{\theta_{\text{ref}}}\left(\boldsymbol{y}|\boldsymbol{x}\right)$~\cite{hsu2020revisiting}.
Similarly, GRPO also suffers from this phenomenon due to the PPO-style hard constraints, and it deteriorates the effective model training.
The analysis in \cite{yu2025dapo} also shows that an upper clipping threshold restricts the probability increase of the ``exploration'' token.
This indicates that the safeguards in GRPO negatively influence model optimization.

\paragraph{Vanishing advantage problem.}
The vanishing advantage problem~\cite{wang2025vl} indicates that the advantage of each sample within the group becomes zero, when the rewards of outputs in the group are equal.
It is problematic since the model cannot receive any signals from the training sample where the advantage is zero for every response.
In particular, we observe that this issue often arises when training samples are either too easy or too difficult for the current model.
Training samples with extreme difficulty levels show worse performance than those with moderate difficulty levels.

\subsection{Regressive GRPO}
Here, we present a \textbf{Reg-GRPO}~(\textbf{Reg}ressive \textbf{G}roup \textbf{R}elative \textbf{P}olicy \textbf{O}ptimization), which reformulates GRPO into the regression task, removing safeguards such as the min and clipping functions.
This reformulation enables the model to directly predict the advantages, resulting in improved alignment of the model with the preference.
Following existing RL-based works~\cite{peters2007reinforcement,peters2010relative,rafailov2023direct,gao2024rebel}, the Reg-GRPO loss function is derived from the RL objective that maximizes the expected reward with the KL constraints between $\pi_\theta$ and $\pi_{\theta_{\text{old}}}$.

\paragraph{RL objective.}
The objective of our reinforcement learning algorithm for each iteration is to maximize rewards while preventing $\pi_\theta$ from making excessive changes relative to $\pi_{\theta_{\text{old}}}$:
\begin{equation}
  \label{eq:objective}
  \pi_{\theta}^* = \underset{\pi_{\theta}}{\mbox{arg}\max}\  \mathbb{E}_{\boldsymbol{x},\boldsymbol{y} \sim \pi_{\theta}\left(\cdot| \boldsymbol{x}\right)} \mathcal{R}\left(\boldsymbol{x}, \boldsymbol{y} \right) - \lambda \; \mathbb{E}_{\boldsymbol{x}}\left[ \mathbb{D}_{\text{KL}}\left(\pi_{\theta}\left(\cdot | \boldsymbol{x} \right) || \pi_{{\theta_{\text{old}}}}\left(\cdot | \boldsymbol{x} \right) \right)\right],
\end{equation}
where $\lambda$ is the hyperparameter that adjusts the strength of the KL-divergence.
Following prior works~\cite{rafailov2023direct, gao2024rebel}, the closed-form solution to the above equation~(Eq.~\eqref{eq:objective}) can be obtained by minimum relative entropy problem:
\begin{equation}
  \label{eq:pi}
  {\pi}_{\theta}^*\left(\boldsymbol{y} | \boldsymbol{x} \right) = \frac{1}{Z\left(\boldsymbol{x}\right)} \pi_{\theta_{\text{old}}}\left(\boldsymbol{y} | \boldsymbol{x} \right)\exp\left(\frac{1}{\lambda} \mathcal{R}\left(\boldsymbol{x}, \boldsymbol{y} \right) \right), \forall \boldsymbol{x}, \boldsymbol{y}
\end{equation}
where $Z\left(\boldsymbol{x}\right) = \sum_{\boldsymbol{y}}{\pi}_{\theta_{\text{old}}} \left(\boldsymbol{y}|\boldsymbol{x} \right) \exp \left(\frac{1}{\lambda}\mathcal{R}\left(\boldsymbol{x}, \boldsymbol{y} \right) \right)$ is a partition function.
However, since calculating the partition function $Z\left(\boldsymbol{x}\right)$ is expensive, it is hard to obtain $\pi_\theta^*$ exactly.

\paragraph{Reg-GRPO Loss.}
To address these issues, we propose \textbf{Reg-GRPO}~(\textbf{Reg}ressive \textbf{GRPO}) loss, which learns the policy model to regress the advantage $\hat{A}^{(i)}$ using the reparameterization, removing the normalization term $Z\left(\boldsymbol{x}\right)$.
Specifically, the advantage for the $i$-th sample is defined as
\begin{equation}
  \label{eq:advantage}
  \hat{A}^{(i)} =  \frac{\mathcal{R}\left(\boldsymbol{x}, \boldsymbol{y}^{(i)} \right) - \mu_r}{\sigma_r},
\end{equation}
where $\mu_r, \sigma_r$ denote the average and standard deviation values of a set of rewards in the group, respectively.
We can also rewrite Eq.~\eqref{eq:pi} to express the reward $\mathcal{R}\left(\boldsymbol{x}, \boldsymbol{y}\right)$ in terms of the optimal model $\pi_\theta^*$, which can be formulated as
\begin{equation}
  \label{eq:reward}
  \mathcal{R}\left(\boldsymbol{x}, \boldsymbol{y} \right) = \lambda \cdot \left(\log Z\left(\boldsymbol{x} \right) + \log \left(\frac{{\pi}_{\theta}^*\left(\boldsymbol{y} | \boldsymbol{x} \right)}{\pi_{\theta_{\text{old}
  }}\left(\boldsymbol{y} | \boldsymbol{x} \right)} \right)\right)\quad \forall \boldsymbol{x},\boldsymbol{y}.
\end{equation}
Since the reward can be expressed through the optimal policy $\pi_\theta^*$ (Eq.~\eqref{eq:reward}), the advantage can be equivalently written as $\hat{A}^{(i)} = \frac{  \rho^*\left(\boldsymbol{x},\boldsymbol{y}^{(i)} \right) -  \mu_{\rho^*} }{\sigma_{\rho^*}}$, where $\rho^*\left(\boldsymbol{x}, \boldsymbol{y}\right)$ is defined as $\rho^*\left(\boldsymbol{x}, \boldsymbol{y}\right)=\log\frac{\pi_{\theta}^*\left(\boldsymbol{y} | \boldsymbol{x} \right)}{\pi_{\theta_{\text{old}}}\left(\boldsymbol{y} | \boldsymbol{x} \right)}$ and $\mu_{\rho^*}, \sigma_{\rho^*}$ denote mean and standard deviation of $\left\{ \rho^*\left(\boldsymbol{x}, \boldsymbol{y}^{(i)} \right)\right\}_{i=1}^{G}$, respectively.
Interestingly, we can see that $Z\left(\boldsymbol{x}\right)$ is naturally removed during the reformulation.

Building on this insight, we define the predictive advantage, which estimates the advantage calculated by normalizing rewards within a group of samples, using the current policy $\pi_\theta$ as
\begin{equation}
  \label{eq:adv_approx}
  \hat{A}^{(i)}_\theta = \frac{  \rho\left(\boldsymbol{x},\boldsymbol{y}^{(i)} \right) -  \mu_{\rho} }{\sigma_{\rho}}, \quad \rho\left(\boldsymbol{x}, \boldsymbol{y} \right) = \log\frac{\pi_{\theta}\left(\boldsymbol{y} | \boldsymbol{x} \right)}{\pi_{\theta_{\text{old}}}\left(\boldsymbol{y} | \boldsymbol{x} \right)},
\end{equation}
where $\mu_{\rho}, \sigma_{\rho}$ are mean and standard deviation of $\left\{ \rho\left(\boldsymbol{x}, \boldsymbol{y}^{(i)} \right)\right\}_{i=1}^{G}$, respectively.
Then, we optimize the policy by minimizing the gap between the target advantage $\hat{A}$ and its predicted counterpart $\hat{A}_\theta$ using Reg-GRPO~(Regressive GRPO), which is defined as:
\begin{equation}
  \mathcal{L}_{\text{Reg-GRPO}}\left(\theta \right) = \mathbb{E}_{\boldsymbol{x}, \left\{\boldsymbol{y}^{(i)} \right\}_{i=1}^G \sim \pi_{\theta_\text{old}}\left(\cdot | \boldsymbol{x}\right)} \left\{\left(\hat{A}^{(i)}- \hat{A}_\theta^{(i)}\right)^2 - \beta \; \mathbb{D}_{\text{KL}}\left[\pi_\theta || \pi_{\text{ref}} \right]\right\},
\end{equation}
Similar to GRPO, we regularize the update with the KL divergence.
The proposed Reg-GRPO loss serves as an effective alternative for optimizing group-level objectives, showing better performance than GRPO.
In our experiments, we demonstrate that this formulation leads to faster convergence and improved policy quality.
\textcolor{black}{The detailed derivation procedure of Reg-GRPO is in Appendix~\ref{app_sec:derivation}.}

\begin{figure*}[t]
  \centering
  \includegraphics[width=1.0\textwidth]{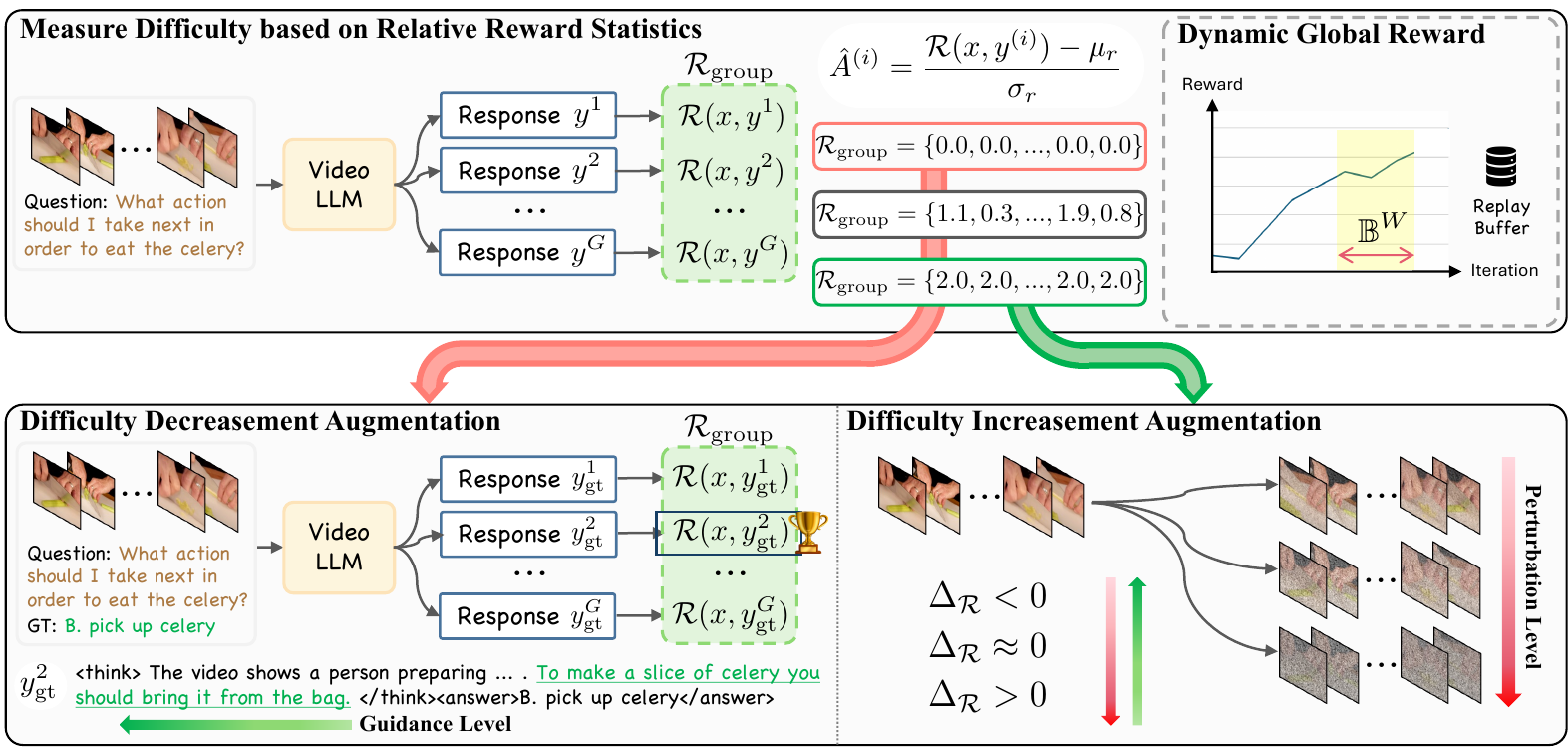}
  \caption{\textbf{Overview of the difficulty-aware data augmentation.}
    First, we assess the difficulty of responses given the input video and question using Eq.~\eqref{eq:difficulty}.
    For hard samples, it augments the input prompts with the reasoning cues extracted from successful reasoning paths~(Difficulty-decreasing augmentation), while the easy samples are perturbed with the noise~(Difficulty-increasing augmentation).
    The scale of the guidance level or noise level is adaptively determined based on the difficulty of the current sample.
  }
  \label{fig:Difficulty}
\end{figure*}

\subsection{Difficulty-aware data augmentation}

In this section, we present a \textbf{difficulty-aware data augmentation} framework, which addresses the vanishing advantage problem in GRPO.
This issue arises when training samples are either too easy or too difficult, leading to uniform rewards across multiple responses.
As a result, the advantage values become zero, erasing the learning signal.
Our augmentation strategy mitigates this issue by modulating the difficulty of inputs to increase variance in predicted rewards, thereby preserving informative gradients for effective model optimization.

Specifically, given an input sample $\boldsymbol{x}=\left(\boldsymbol{v}, \boldsymbol{q} \right)$, where $\boldsymbol{v}$ and $\boldsymbol{q}$ denote a video and a question, respectively, we first generate multiple responses and then compute the average reward $\frac{1}{G}\sum_{i=1}^G\mathcal{R}\left(\boldsymbol{x}, \boldsymbol{y}^{(i)} \right)$ for the sample.
We then measure the difficulty of $\boldsymbol{x}$ by comparing its average reward with the average reward of samples in a replay buffer $\mathbb{B}^{W}$, which consists of samples $\boldsymbol{x}_{\text{rep}}$ and their corresponding outputs $\left\{\boldsymbol{y}^{(i)}_{\text{rep}}\right\}_{i=1}^G$ from the most recent $W$ steps.
Formally, the difficulty $\Delta_{\mathcal{R}}\left(\boldsymbol{x} \right)$ of the input sample $\boldsymbol{x}$ is calculated as:
\begin{equation}
  \label{eq:difficulty}
  \Delta_{\mathcal{R}}\left(\boldsymbol{x}\right) = \mathbb{E}_{\left(\boldsymbol{x}_{\text{rep}}, \left\{\boldsymbol{y}_{\text{rep}}^{(i)}\right\}_{i=1}^G\right) \in \mathbb{B}^{W}}\left[\frac{1}{G}\sum_{i=1}^G\mathcal{R}\left(\boldsymbol{x}_{\text{rep}}, \boldsymbol{y}_{\text{rep}}^{(i)} \right)\right] - \frac{1}{G}\sum_{j=1}^G\mathcal{R}\left(\boldsymbol{x}, \boldsymbol{y}^{(j)}\right).
\end{equation}
Instead of using only the sample reward $\mathcal{R}\left(\boldsymbol{x},\boldsymbol{y}\right)$, we use the average reward of samples in the replay buffer as a reference value to account for model evolution.

Building on the difficulty metric in Eq.~\eqref{eq:difficulty}, we adaptively adjust each training sample to balance the learning signal.
For samples identified as too easy (\ie high reward and low difficulty), we increase task complexity by perturbing the input---for example, by adding Gaussian noise to video inputs or masking video frames---to encourage the model to attend to more informative cues.
Conversely, for too difficult samples (\ie low reward and high difficulty), we inject auxiliary reasoning hints or visual cues that simplify the temporal context, helping the model focus on core reasoning paths instead of failing due to overwhelming difficulty.
This dynamic modulation leads to an appropriate difficulty distribution across training, preventing the collapse of advantage values while ensuring that each update provides a meaningful gradient signal.

\paragraph{Difficulty-decreasing augmentation.}
For difficult samples~($\Delta_{\mathcal{R}}\left(\boldsymbol{x} \right)>0$), we ease the difficulty of the sample by providing auxiliary reasoning cues that guide the model toward generating correct reasoning.
Concretely, we first augment the input prompt $\boldsymbol{q}$ with the ground-truth answer and generate multiple reasoning trajectories using the VideoLLM.
Among these reasoning trajectories, we select the response $\boldsymbol{y}_{\text{gt}}$ with the highest reward and extract a partial reasoning trace of it.
This trace is then incorporated into the original prompt to form a modified prompt $\tilde{\boldsymbol{q}}$ containing structured hints that guide the model's reasoning toward the correct solution.
To maintain adaptive control, the guidance level is scaled according to the sample’s difficulty magnitude.
Harder samples receive stronger reasoning cues, while moderately difficult samples are given lighter guidance.
By adaptively providing guidance for challenging inputs, this augmentation mitigates vanishing gradients in difficult cases and facilitates more stable convergence through progressively refined reasoning.

\paragraph{Difficulty-increasing augmentation.}
Conversely, for easy samples~($\Delta_{\mathcal{R}}\left(\boldsymbol{x} \right)<0$), we employ a difficulty-increasing augmentation to enhance task complexity and encourage the model to explore more diverse reasoning trajectories.
We perturb the visual input $\boldsymbol{v}$ to create a harder input $\tilde{\boldsymbol{v}}$ by introducing frame-level Gaussian noise, thereby slightly degrading perceptual fidelity while preserving overall semantic structure.
The intensity of the noise is proportionally scaled by the difficulty magnitude, ensuring that easier samples receive stronger perturbations and moderately easy samples remain stable.
This adaptive corruption expands the diversity of generated reasoning trajectories.
By inducing distributed rewards, the augmentation ensures that even trivial samples provide informative gradients, avoiding zero learning signals throughout the optimization process.

{
\begin{table}[t]
    \caption{\textbf{Performance on SEED-Bench-R1 validation split and LongVideoBench.} In-Dist. means in-distribution dataset.}
    \label{tab:main_sbr}
    \centering
    \renewcommand{\arraystretch}{1.1}
    \begin{adjustbox}{width=1.0\textwidth}
    \begin{tabular}{lcccccccccccc}
        \toprule
        \multirow{3}{*}{{\textbf{Method}}} & SBR-L1  & SBR-L2 & \multicolumn{5}{c}{SBR-L3} & \multicolumn{5}{c}{LongVideoBench}\\
         & In-Dist. & Cross-Env & \multicolumn{5}{c}{Cross-Task, Cross-Env} & \multicolumn{5}{c}{Cross-Task, Cross-Env}\\
        \cmidrule(lr){2-2} \cmidrule(lr){3-3} \cmidrule(lr){4-8} \cmidrule(lr){9-13}
          & Daily life & Daily life & Daily life & Hobbies & Recreation & Work & Overall & (8,15] & (15,60] & (180,600] & (900,3600] & Overall \\
        \hline
        \rowcolor{gray!10}
        \multicolumn{13}{l}{\textbf{\textit{Baseline}}} \\
        VideoLLaMA3-7B~\cite{zhang2025videollama}   & 33.3 & 33.2 & 26.7 & 28.5 & 30.6 & 27.0 & 27.7 &35.7 & 43.1 & 21.0 & 22.5 & 26.7 \\
        InternVL3-2B~\cite{zhu2025internvl3}        & 23.7 & 23.1 & 21.2 & 16.3 & 18.6 & 12.6 & 17.1 & 41.6 & 48.4 & 33.7 & 30.0 & 34.8 \\
        InternVL3-8B~\cite{zhu2025internvl3}        & 41.4 & 40.8 & 39.2 & 34.6 & 35.0 & 30.0 & 34.8 & 54.6 & 66.7 & 46.1 & 44.2 & 49.1\\
        InternVL3-14B~\cite{zhu2025internvl3}       & 43.6 & 45.8 & 44.0 & 34.9 & 36.6 & 31.7 & 37.2 & 69.2 & 62.8 & 46.3 & 42.0 & 50.0 \\
        \hline
        \rowcolor{black!20}
        \multicolumn{13}{l}{\textbf{\textit{Qwen2-VL-2B}}} \\
        Qwen2-VL-2B~\cite{wang2024qwen2}   & 12.9 & 16.4 & 13.0 & 16.3 & 10.4 & 14.4 & 13.8 & 33.0 & 32.7 & 29.5 & 22.2 & 27.2 \\
        SFT                   & 34.1 & 36.2 & 36.9	& 34.6 & 30.0 & 33.9 & 33.8 & 42.6 & 42.3 & 37.0 & 32.2 & 36.3\\
        GRPO                                & 38.4 & 42.0 & 40.6 & 37.6 & 30.5 & 40.4 & 36.8 & 47.0 & 46.4 & 36.6 & 32.1 & 37.4 \\
        \rowcolor{Cyan!20}
        \textbf{DeepVideo-R1} 
        & \textbf{48.9} & \textbf{50.3} & \textbf{52.4} & \textbf{42.7} & \textbf{41.1} & \textbf{49.2} & \textbf{46.3} & \textbf{51.4} & \textbf{49.7} & \textbf{38.5} & \textbf{34.8} & \textbf{40.1} \\
        \hline
        \rowcolor{black!20}
        \multicolumn{13}{l}{\textbf{\textit{Qwen2-VL-7B}}} \\
        Qwen2-VL-7B~\cite{wang2024qwen2}   & 34.8 & 34.0 & 31.2 & 32.3 & 33.3 & 30.7 & 31.6 & 42.3 & 44.8 & 34.0 & 25.4 & 32.9 \\
        SFT                   & 43.8 & 44.1 & 38.3 & 41.0 & 32.2 & 38.6 & 38.2 & 45.0 & 54.7 & 36.7	& 36.4 & 40.0 \\
        GRPO                                & 46.0 & 50.2 & 48.5 & 45.1 & 43.7 & 41.3 & 44.9 & 54.5 & 53.5 & 42.5 & 37.2 & 43.4 \\
        \rowcolor{Cyan!20}
        \textbf{DeepVideo-R1} 
        & \textbf{56.4} & \textbf{59.8} & \textbf{57.6} & \textbf{52.5} & \textbf{50.0} & \textbf{55.2} & \textbf{53.8} & \textbf{56.2} & \textbf{61.4} & \textbf{45.1} & \textbf{40.8} & \textbf{46.9} \\
        \hline
        \rowcolor{black!20}
        \multicolumn{13}{l}{\textbf{\textit{Qwen2.5-VL-3B}}} \\
        Qwen2.5-VL-3B~\cite{bai2025qwen2}   & 31.3 & 32.7 & 33.0	& 28.8 &	27.3&	23.0&	28.2  & 50.3 & 62.1 & 39.0 & 32.1 & 40.4 \\
        SFT                   & 35.9 & 39.1 & 39.9 & 31.5 & 29.7 & 31.2 & 33.7 & 51.4 &	52.9 & 36.3 & 35.3 & 39.7 \\
        GRPO                                & 39.6 & 41.0 & 39.0 & 33.9 & 36.6 & 31.9 & 35.4 & 51.4 & 62.1 & 42.2 & 36.0 & 43.2 \\
        \rowcolor{Cyan!20}
        \textbf{DeepVideo-R1} 
        & \textbf{48.1} & \textbf{51.1} & \textbf{46.5} & \textbf{45.8} & \textbf{43.7} & \textbf{40.1} & \textbf{44.0} & \textbf{54.1} & \textbf{64.1} &	\textbf{45.9} & \textbf{43.4} &	\textbf{48.4} \\
        \hline
        \rowcolor{black!20}
        \multicolumn{13}{l}{\textbf{\textit{Qwen2.5-VL-7B}}} \\
        Qwen2.5-VL-7B~\cite{bai2025qwen2}               & 33.4 & 38.2 & 35.1 & 31.5 & 27.3 & 28.0 & 31.0  & 54.6	& \textbf{63.4} & 37.8	& 36.2 & 42.5 \\
        SFT                             & 42.4 & 42.6 & 41.2 & 37.3 & 36.6 & 41.5 & 39.0 & 54.6 & 56.2 & 41.5 & 38.1 & 43.9 \\
        GRPO   & 49.1 & 52.1 & 49.4 & 40.7 & 43.2 & 35.2 & 42.2 & 61.1 & 60.8 & 44.4 & 40.8 & 47.7 \\
        \rowcolor{Cyan!20}
        \textbf{DeepVideo-R1}
        &\textbf{52.0} & \textbf{55.7} & \textbf{51.3} & \textbf{47.8} & \textbf{47.0} & \textbf{44.1} & \textbf{47.7} & \textbf{62.7} & \textbf{63.4} & \textbf{49.3} & \textbf{44.5} & \textbf{51.1} \\
        \hline
    \end{tabular}
    \end{adjustbox}
\end{table}
}

\section{Experiments}

\subsection{Experimental Settings}
To validate the effectiveness of the proposed method, we conduct evaluations on various video benchmarks, including both general video understanding tasks (\eg SEED-Bench-R1~\cite{chen2025exploring}, VSI-Bench, Video-MMMU, MMVU (mc), MVBench, TempCompass, Video-MME (wo sub)), long video understanding tasks~(\eg LongVideoBench~\cite{wu2024longvideobench}), and fine-grained spatial-temporal video reasoning tasks (NExTGQA~\cite{xiao2024can}).
More details about datasets are in Appendix~\ref{app_sec:datasets}.
We employ Qwen2-VL-2B/7B~\cite{wang2024qwen2} and Qwen2.5-VL-3B/7B~\cite{bai2025qwen2} for the experiments.
For the analysis, we use Qwen2.5-VL-3B as a default video LLM.
More implementation details are in Appendix~\ref{app_sec:impl_details}.

\subsection{Experimental Results}
\paragraph{Experimental results on SEED-Bench-R1.}
Table \ref{tab:main_sbr} summarizes the performance of various baselines, supervised fine-tuning (SFT), GRPO, and our proposed DeepVideo-R1 on the validation splits of the SEED-Bench-R1 (SBR) dataset.
Across all settings (SBR-L1, L2, L3), DeepVideo-R1 consistently achieves the best performance, demonstrating its strong capability for video reasoning under both in-distribution and cross-environment settings.
Specifically, compared with Qwen2.5-VL-3B + GRPO, our DeepVideo-R1-3B improves the overall scores on SBR-L1, L2, and L3 by +8.5, +10.1, and +8.6 points, respectively.
Notably, the performance gains on SBR-L2 and L3 (overall) exceed those on SBR-L1, indicating that DeepVideo-R1 enhances generalization across cross-task and cross-environment settings.
These results suggest that the proposed regression-based optimization and reasoning-aware augmentation in DeepVideo-R1 enable more stable policy learning and improved adaptability to diverse video understanding scenarios.

\paragraph{Experimental results on LongVideoBench.}
We further evaluate DeepVideo-R1 on LongVideoBench, a benchmark designed to assess long-video reasoning and temporal compositional understanding.
As shown in Table \ref{tab:main_sbr}, DeepVideo-R1 again outperforms all baselines across varying temporal ranges, achieving an overall score of 51.1, surpassing both SFT and GRPO-trained counterparts.
In particular, DeepVideo-R1-3B achieves a substantial improvement of +7.4 over Qwen2.5-VL-3B + GRPO on the longest duration range (900 – 3600 s), underscoring its superior ability to reason over extended temporal contexts.
This strong performance on long-duration videos highlights DeepVideo-R1’s effectiveness in maintaining coherent reasoning over time, validating its robustness in complex real-world video understanding tasks.

{
\begin{table}[t]
    \caption{\textbf{Performance on various video reasoning and general benchmarks.}}
    \label{tab:video-r1}
    \centering
    \renewcommand{\arraystretch}{1.1}
    \begin{adjustbox}{width=1.0\textwidth}
    \begin{tabular}{lccc|ccc}
        \toprule
        \multirow{2}{*}{{\textbf{Method}}}   & \multicolumn{3}{c}{Video Reasoning Benchmark} & \multicolumn{3}{c}{Video General Benchmark} \\
        \cmidrule(lr){2-7}
        & VSI-Bench & Video-MMMU & MMVU~(mc) & MVBench & TempCompass & Video-MME~(wo sub) \\
        \hline
        LLaMA-VID~\cite{li2024llama}                 & -    & -    & -    & 41.9 & 45.6 & -    \\
        VideoLLaMA2~\cite{cheng2024videollama}       & -    & -    & 44.8 & 54.6 & -    & 47.9 \\
        LongVA-7B~\cite{zhang2024long}               & 29.2 & 23.9 & -    & -    & 56.9 & 52.6 \\
         VILA-1.5-8B~\cite{lin2024vila}               & 28.9 & 20.8 & -    & -    & 58.8 & -    \\
        VILA-1.5-40B~\cite{lin2024vila}              & 31.2 & 34.0 & -    & -    & -    & 60.1 \\
        Video-UTR-7B~\cite{yu2025unhackable}         & -    & -    & -    & 58.8 & 59.7 & 52.6 \\
        LLaVA-OneVision-7B~\cite{li2024llava}        & 32.4 & 33.8 & 49.2 & 56.7 & -    & 58.2 \\
        Kangaroo-8B~\cite{liu2024kangaroo}           & -    & -    & -    & 61.1 & 62.5 & 56.0 \\
        \hline
        Qwen2.5-VL-3B~\cite{bai2025qwen2}            & 32.4 & 36.1 & 54.2 & 48.1 & 29.7 & \textbf{54.4} \\
        \rowcolor{Cyan!20}
        \textbf{DeepVideo-R1-3B~(Ours)}   & \textbf{33.0} & \textbf{40.7} & \textbf{59.0} & \textbf{49.6} & \textbf{63.1} & 51.1 \\
        \hline
    \end{tabular}
    \end{adjustbox}
\end{table}
}
\begin{table}[t]
\begin{minipage}[t!]{0.49\textwidth}
    \caption{\textbf{Experimental results on NExTGQA}}
    \label{tab:table_gqa}
    \centering
    \renewcommand{\arraystretch}{1.1}
    \setlength{\tabcolsep}{8pt}
    \footnotesize
    \resizebox{\textwidth}{!}{
    \begin{tabular}{lcc}
         \toprule
        {{\textbf{Method}}} & mIoU & Acc@QA \\
        \hline
        \rowcolor{gray!10}
        \multicolumn{3}{l}{\textbf{\textit{Vision Experts}}} \\
          IGV~\cite{li2022invariant} & 14.0 & 50.1 \\
          Temp[CLIP]~\cite{xiao2024can} & 12.1 & 60.2 \\
          FrozenBiLM~\cite{yang2022zero} & 9.6 & 70.8   \\
          SeViLA~\cite{yu2023self} & 21.7 & 68.1  \\
          \hline
        \rowcolor{gray!10}
        \multicolumn{3}{l}{\textbf{\textit{VideoLLMs}}} \\
          VideoChat-R1~\cite{li2025videochat}     & 32.4 & 70.6 \\
          VideoChat-R1-thinking~\cite{li2025videochat}     & 36.1 & 69.2 \\
          \hline
         
         \rowcolor{Cyan!20}
         \textbf{DeepVideo-R1-7B (Ours)}     & \textbf{36.8} & \textbf{72.5}  \\
         \hline
    \end{tabular}
}
\end{minipage}
\hfill
\begin{minipage}[t!]{0.50\textwidth}
    \caption{\textbf{Ablation study} on training schemes~(Reg-GRPO and difficulty-aware data augmentation~(DA-Aug.) in DeepVideo-R1 using SEED-Bench-R1 dataset.}
    \label{tab:abl_table_main}
    \centering
    \renewcommand{\arraystretch}{1.1}
    \setlength{\tabcolsep}{8pt}
    \footnotesize
    \resizebox{\textwidth}{!}{
    \begin{tabular}{cc|ccc}
         \toprule
         Method & DA-Aug. & L1~(In-Dist.) & L2~(OOD) & L3~(OOD) \\
         \midrule
          Qwen2.5-VL-3B &                        & 31.3 & 32.7 & 27.0  \\
          \midrule
         GRPO  & & 39.6 & 41.0 & 35.4  \\
          GRPO     & \checkmark & 41.7 & 42.5 & 36.6  \\
          \midrule
         Reg-GRPO &                & 44.2 & 44.2 & 39.5  \\
         \rowcolor{Cyan!20}
         Reg-GRPO  &\checkmark          & \textbf{48.1} & \textbf{51.1} & \textbf{44.0}  \\
         \bottomrule
    \end{tabular}
    }
\end{minipage}
\end{table}

\paragraph{Experimental results on various video benchmarks.}
We further evaluate DeepVideo-R1-3B, built upon Qwen2.5-VL-3B, on diverse video reasoning and general benchmarks to assess its generalization ability.
Following \cite{feng2025video}, we use the Video-R1~\cite{feng2025video} training set to train the model, and report the results in Table~\ref{tab:video-r1}.
As shown in the table, DeepVideo-R1-3B consistently outperforms Qwen2.5-VL-3B and other large-scale multimodal video models (e.g., LLAVA-OneVision-7B, VILA-1.5-40B, Kangaroo-8B) across almost all benchmarks.
In comparison of the experimental results of the base model Qwen2.5-VL-7B, our DeepVideo-R1 consistently improves the performance on 5 out of 6 datasets.
In particular, DeepVideo-R1 improves performance from 29.7 to 63.1 on TempCompass.
Overall, these results confirm that DeepVideo-R1 effectively generalizes beyond SEED-Bench-R1, establishing a new performance level across both reasoning-oriented and general video understanding benchmarks.

\paragraph{Experimental results on NExTGQA.}
Table \ref{tab:table_gqa} reports the performance of DeepVideo-R1 compared with both vision experts (IGV, Temp[CLIP], FrozenBiLM, SeViLA) and VideoLLMs (VideoChat-R1, VideoChat-R1-thinking).
For the NExT-GQA benchmark, DeepVideo-R1-7B is trained with a composite reward combining accuracy, format consistency, and IoU, aligning with the dataset’s grounding-oriented evaluation.
As shown, DeepVideo-R1-7B achieves 36.8 mIoU and 72.5 Acc@QA, outperforming all baselines, including a +4.4 mIoU and +2.3 Acc@QA gain over VideoChat-R1.
These improvements highlight DeepVideo-R1’s ability to enhance both spatial grounding and reasoning precision, demonstrating its robustness to diverse reward designs and its strong adaptability across grounded video reasoning tasks.

\subsection{Analysis}

\paragraph{Ablation studies.}
We conduct ablation studies to explore the contribution of Reg-GRPO and difficulty-aware data augmentation~(DA-Aug.) in Table~\ref{tab:abl_table_main}.
The table demonstrates that both Reg-GRPO and difficulty-aware data augmentation contribute to the performance improvement of DeepVideo-R1.
By comparing the GRPO-trained model without and with difficulty-aware data augmentation, we observe a +2.1-point improvement on SBR~(L1), suggesting that difficulty-aware data augmentation is effective for both GRPO and Reg-GRPO.
Also, Reg-GRPO~(w/o DA-Aug.) yields a +4.1-point improvement over Qwen2.5-VL-3B+GRPO~(w/o DA-Aug.) on SBR~(L3).
This indicates that Reg-GRPO is more effective than GRPO by directly predicting advantages.

\begin{table}
  \begin{minipage}[t!]{0.52\textwidth}
    \caption{ Performance comparison on reinforcement learning algorithm.}
    \label{tab:analysis_on_RFT}
    \centering
    \renewcommand{\arraystretch}{1.1}
    \setlength{\tabcolsep}{16pt}
    \footnotesize
    \resizebox{\textwidth}{!}{
      \begin{tabular}{l|ccc}
        \toprule
        Method & L1 & L2 & L3 \\
        \midrule
        Qwen2.5-VL-3B                  & 31.3 & 32.7 & 27.0  \\
        \midrule
        + DPO~\cite{rafailov2023direct}  & 35.8 & 35.2 & 30.8  \\
        + Online DPO~\cite{rafailov2023direct} & 37.1 & 38.1 & 31.9  \\
        + REINFORCE~\cite{williams1992simple,kreutzer2017bandit,nguyen2017reinforcement}                     & 37.0 & 39.5 & 32.3  \\
        + RLOO~\cite{kool2019buy}                       & 35.0 & 37.4 & 31.3  \\
        + REBEL~\cite{gao2024rebel}                          & 41.8 & 43.7 & 38.0  \\
        + Reward-Regression               & 32.5 & 33.1  & 28.3 \\
        + GRPO~\cite{shao2024deepseekmath,guo2025deepseek}                           & 39.6 & 41.0 &  35.4  \\
        \rowcolor{Cyan!20}
        \textbf{+ Reg-GRPO~(Ours)}                           & \textbf{44.2} & \textbf{44.2} & \textbf{39.5}  \\
        \bottomrule
      \end{tabular}
    }
    \caption{Performance comparison on absolute and relative difficulty measurement.}
    \label{tab:difficulty_ref}
    \centering
    \resizebox{0.85\textwidth}{!}{
      \begin{tabular}{c|ccc}
        \toprule
        Diff. ref.  & L1 & L2 & L3 \\
        \midrule
        Absolute  &  47.9 & 50.1 & 40.7  \\
        \rowcolor{Cyan!20}
        Relative  &   \textbf{48.1} & \textbf{51.1} & \textbf{44.0} \\
        \bottomrule
      \end{tabular}
    }
  \end{minipage}
  \hfill
  \begin{minipage}[t!]{0.42\textwidth}
    \caption{Performance comparison according to the data augmentation types. Diff$\uparrow$ indicates difficulty-increasing augmentation. Diff$\downarrow$ indicates difficulty-decreasing augmentation.}
    \label{tab:abl_augmentation_type}
    \centering
    \renewcommand{\arraystretch}{1.1}
    \setlength{\tabcolsep}{12pt}
    \resizebox{\textwidth}{!}{
      \begin{tabular}{cc|ccc}
        \toprule
        Diff. $\uparrow$ & Diff. $\downarrow$ & L1 & L2 & L3 \\
        \midrule
        &  & 44.2 & 44.2 & 36.3  \\
        \checkmark &    & 45.3 & 46.9 & 40.0  \\
        & \checkmark    & 45.3 & 47.3 & 41.6  \\
        \rowcolor{Cyan!20}
        \checkmark & \checkmark                                 & \textbf{48.1} & \textbf{51.1} & \textbf{44.0}  \\
        \bottomrule
      \end{tabular}
    }
    \caption{Performance comparison on augmentation scaling scheme (Fixed guidance/noise level v.s. Adaptive guidance/noise level).}
    \label{tab:scaling}
    \centering
    \renewcommand{\arraystretch}{1.1}
    \setlength{\tabcolsep}{12pt}
    \footnotesize
    \resizebox{\textwidth}{!}{
      \begin{tabular}{c|ccc}
        \toprule
        Aug.  & L1 & L2 & L3 \\
        \midrule
        No Aug.  & 44.2 & 44.2 & 36.3  \\
        Fixed  &  46.8 & 48.6 & 43.0  \\
        \rowcolor{Cyan!20}
        Adaptive  &   \textbf{48.1} & \textbf{51.1} & \textbf{44.0} \\
        \bottomrule
      \end{tabular}
    }
  \end{minipage}
\end{table}

\begin{figure}[t!]
\centering
\vspace{-12pt}
\begin{subcaptiongroup}
\begin{subfigure}{0.48\textwidth}
    \includegraphics[width=\linewidth]{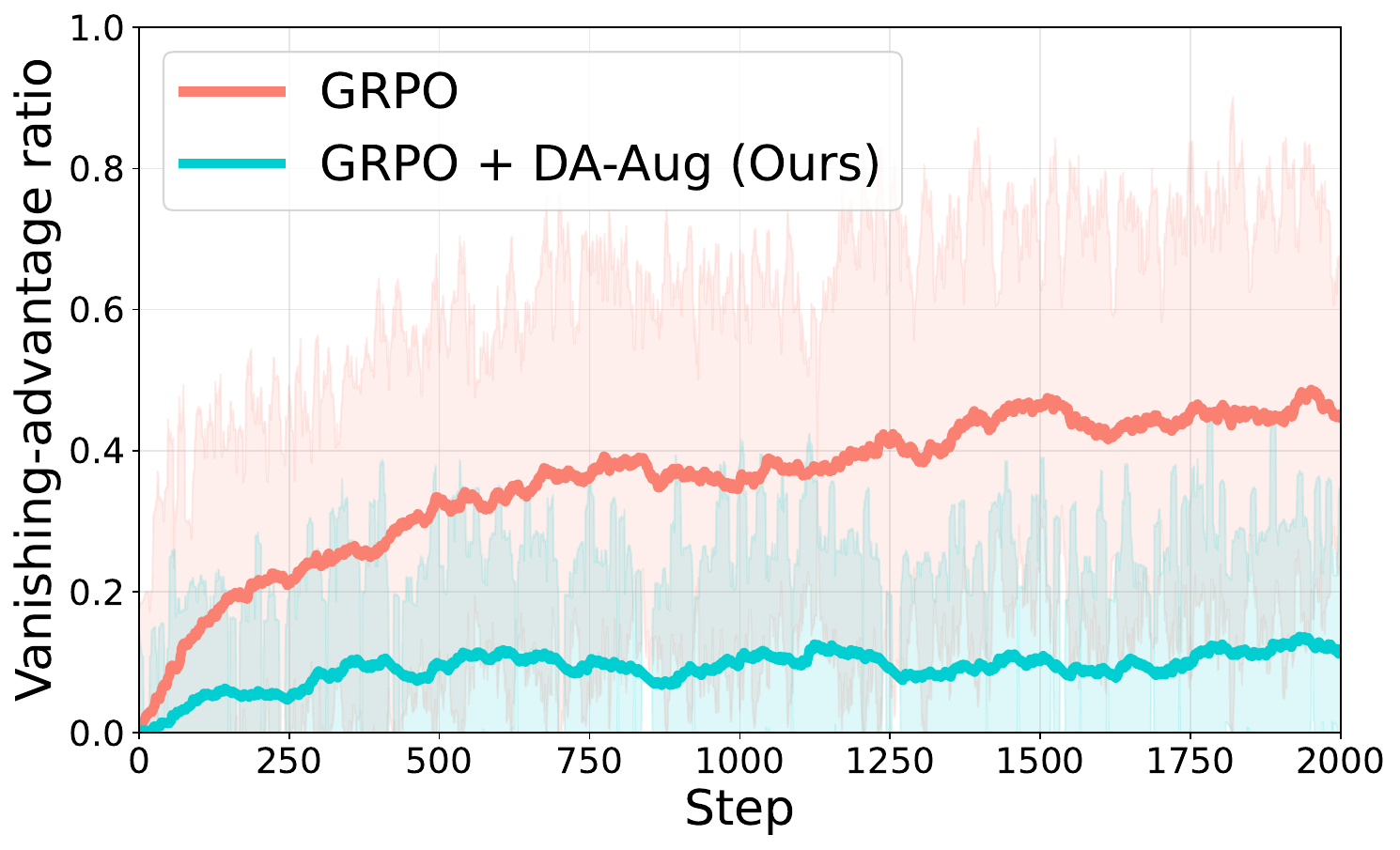}
\end{subfigure}
\begin{subfigure}{0.48\textwidth}
    \includegraphics[width=\linewidth]{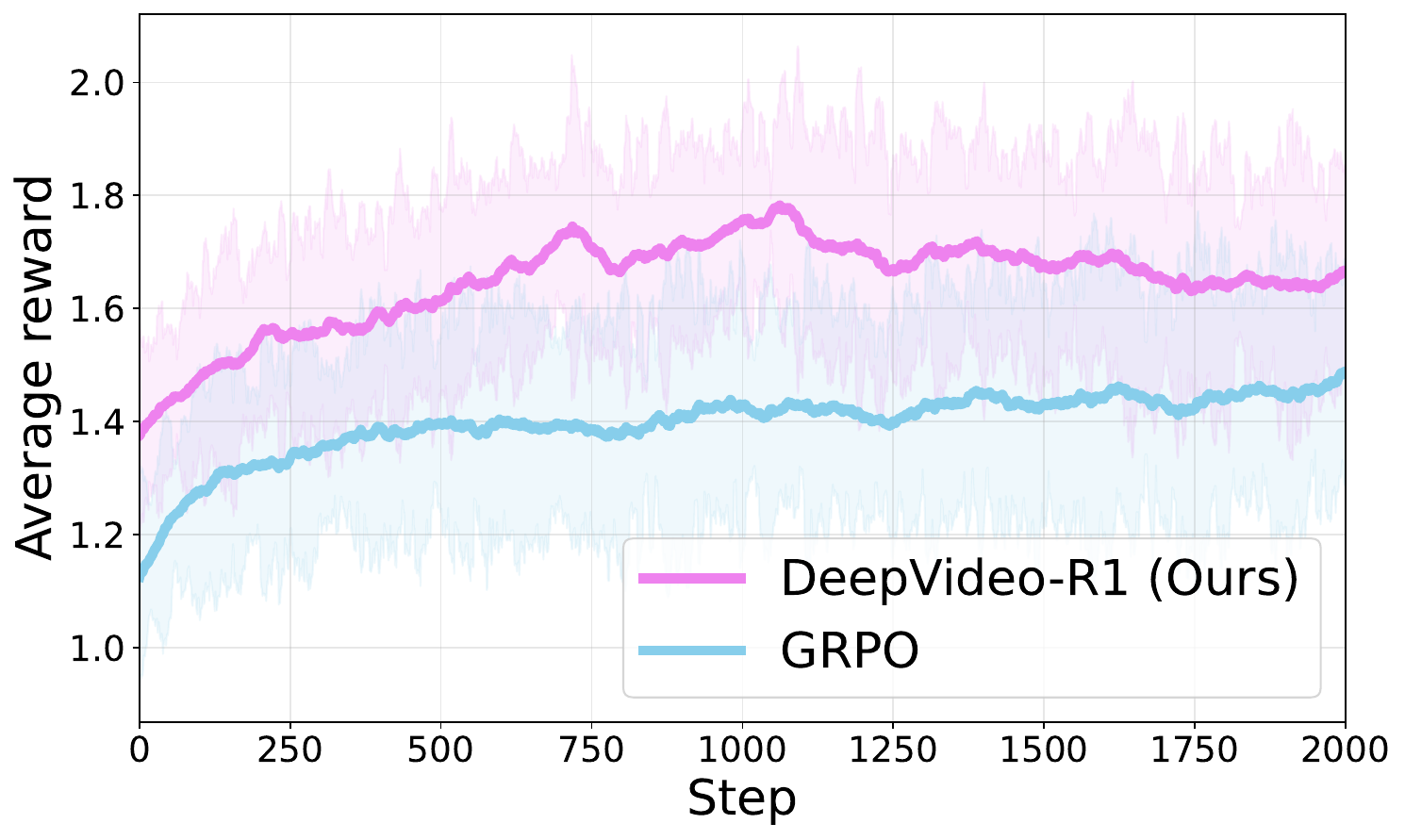}
\end{subfigure}
\end{subcaptiongroup}
\vspace{-6pt}
\caption{
\textbf{Vanishing advantage ratio comparison} on GRPO and GRPO+DA-Aug~(Difficulty-aware augmentation)~(Left). \textbf{Reward curves} of DeepVideo-R1~(Ours) and GRPO~(Right).
}
\label{fig:training_analysis}
\vspace{-8pt}
\end{figure}

\paragraph{Comparison with reinforcement learning methods.}
Table~\ref{tab:analysis_on_RFT} compares our Reg-GRPO with representative reinforcement fine-tuning (RFT) methods.
The detailed explanations about the reinforcement fine-tuning methods are in Appendix~\ref{app_sec:baselines}.
From the table, our DeepVideo-R1 achieves the best performance among the compared RFT methods.
In particular, compared to reward regression, which directly regresses reward scores, DeepVideo-R1 performs significantly better.
This result highlights that directly aligning advantages is more effective than reward regression.

\paragraph{Impact of relative difficulty measurement.}
We compare absolute and relative difficulty measurements for adaptive data augmentation in Table~\ref{tab:difficulty_ref}.
The relative scheme, which considers reward history statistics, consistently outperforms the absolute counterpart across all difficulty levels (L1--L3) on the SEED-Bench-R1 dataset, demonstrating its superior ability to guide effective augmentation.

\paragraph{Impact of difficulty-decreasing/increasing augmentation.}
In Table~\ref{tab:abl_augmentation_type}, we compare different data augmentation schemes: difficulty-increasing and difficulty-decreasing augmentations.
The table demonstrates that the model trained with both schemes achieves the best performance, yielding a +7.7-point gain on SBR~(L3), which is an out-of-distribution dataset.
This reveals that adjusting the sample's difficulty to a moderate level is important for learning with group-normalized advantages.

\paragraph{Impact of augmentation scaling scheme.}
In Table~\ref{tab:scaling}, we compare different augmentation scaling strategies: no augmentation, fixed scaling~(constant guidance/noise level), adaptive scaling~(difficulty-aware guidance/noise).
From the table, the adaptive strategy consistently outperforms the others across all difficulty levels (L1–L3), demonstrating the importance of tailoring augmentation strength based on input difficulty.

\begin{figure*}[t!]
\centering
\includegraphics[width=1.0\textwidth]{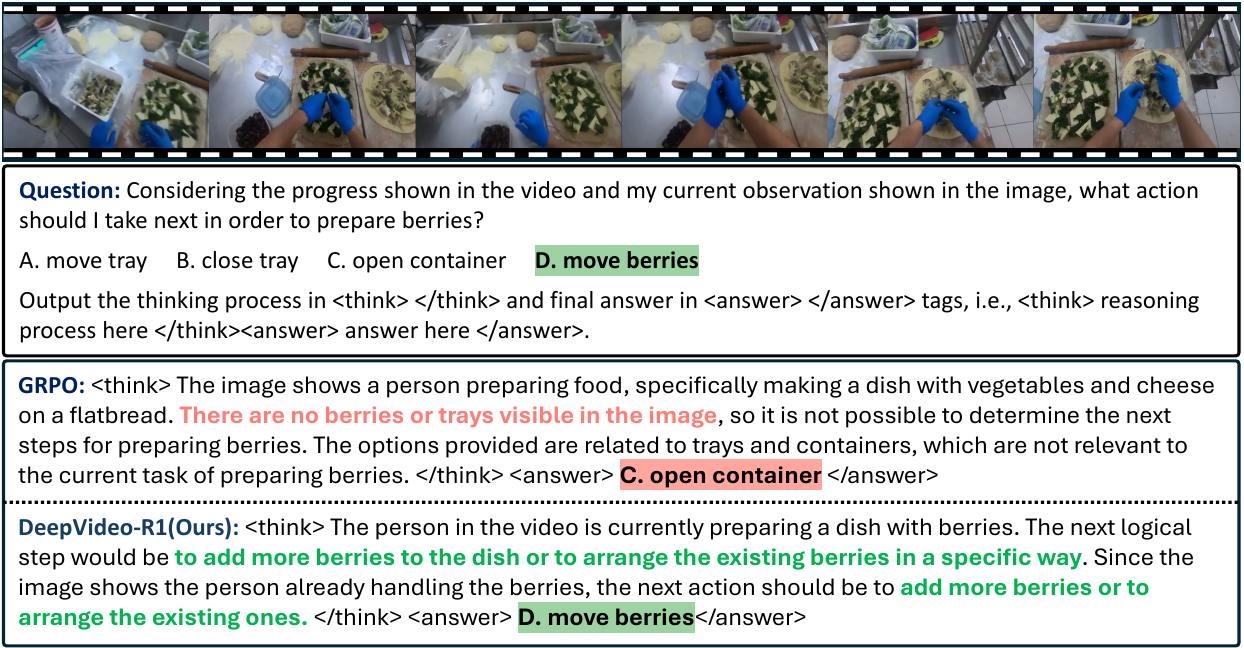}
\caption{\textbf{Qualitative result of DeepVideo-R1-7B} in comparison of Qwen2.5-VL-7B+GRPO.}
\label{fig:qualitative_results}
\end{figure*}

\paragraph{Impact of augmentation.}
We compare the vanishing advantage ratio of GRPO and GRPO with DA-Aug~(difficulty-aware data augmentation) in Figure~\ref{fig:training_analysis}~(left).
The figure shows that our data augmentation effectively reduces the ratio of samples causing the vanishing advantage.
This indicates that the data augmentation effectively addresses the vanishing advantage problem in GRPO.

\paragraph{Reward curves.}
In addition, we plot the reward curves of GRPO and our DeepVideo-R1, where the x-axis is training step and y-axis is the average reward in Figure~\ref{fig:training_analysis}~(right).
From the figure, our DeepVideo-R1 gets a higher average reward with Reg-GRPO and difficulty-aware data augmentation.

\paragraph{Qualitative results.}
Figure~\ref{fig:qualitative_results} presents a qualitative example from SEED-Bench-R1-7B, comparing the outputs of our DeepVideo-R1 and the Qwen2.5-VL-7B trained with GRPO.
The task is to predict the next action given a video.
Our DeepVideo-R1 correctly infers that the person will continue moving berries.
While the GRPO-only model fails to recognize the presence of berries.
This demonstrates that DeepVideo-R1 has strong visual grounding and understanding capabilities.

\section{Conclusion}
We propose a video large language model, DeepVideo-R1, trained with Reg-GRPO~(Regressive GRPO) and difficulty-aware data augmentation to address two key challenges in group relative policy optimization.
Reg-GRPO reformulates the GRPO objective as a regression task that directly aligns the model with group-normalized advantages.
Difficulty-aware data augmentation modulates input difficulty to alleviate the vanishing advantage problem.
Our experiments demonstrate that DeepVideo-R1 consistently improves the reasoning performance of diverse VideoLLMs, outperforming GRPO-based reinforcement fine-tuning.

\section*{Acknowledgment}

This work partly supported by Korea Research Institute for defense Technology planning and advancement - Grant funded by Defense Acquisition Program Administration(DAPA)(KRIT-CT-23-021, 30\%), the InnoCORE program of the Ministry of Science and ICT(N10250156, 30\%), Virtual Engineering Platform Project (Grant No. P0022336, 30\%), funded by the Ministry of Trade, Industry \& Energy (MoTIE, South Korea), and Electronics and Telecommunications Research Institute(ETRI) grant funded by the Korean government [25ZB1200, Fundamental Technology Research for Human-Centric Autonomous Intelligent Systems, 10\%].

{
  \small
  \bibliographystyle{unsrt}
  \bibliography{main}

@String(IJCV = {Int. J. Comput. Vis.})

@String(CVPR= {IEEE Conf. Comput. Vis. Pattern Recog.})

@String(ICCV= {Int. Conf. Comput. Vis.})

@String(ECCV= {Eur. Conf. Comput. Vis.})

@String(ACMMM= {ACM Int. Conf. Multimedia})

@String(ICLR = {Int. Conf. Learn. Represent.})

@String(AAAI = {AAAI})

@String(IJCV  = {IJCV})

@String(CVPR  = {CVPR})

@String(ICCV  = {ICCV})

@String(ECCV  = {ECCV})

@String(ACMMM = {ACM MM})

@String(ICLR  = {ICLR})

@inproceedings{zhu2024self,
  title={Self-supervised visual preference alignment},
  author={Zhu, Ke and Zhao, Liang and Ge, Zheng and Zhang, Xiangyu},
  booktitle={ACMMM},
  year={2024}
}

@inproceedings{rafailov2023direct,
  title={Direct preference optimization: Your language model is secretly a reward model},
  author={Rafailov, Rafael and Sharma, Archit and Mitchell, Eric and Manning, Christopher D and Ermon, Stefano and Finn, Chelsea},
  booktitle={NeurIPS},
  year={2023}
}

@article{gao2024rebel,
    title={REBEL: Reinforcement Learning via Regressing Relative Rewards}, 
    author={Zhaolin Gao and Jonathan D. Chang and Wenhao Zhan and Owen Oertell and Gokul Swamy and Kianté Brantley and Thorsten Joachims and J. Andrew Bagnell and Jason D. Lee and Wen Sun},
    journal={NeurIPS},      
    year={2024},
}

@inproceedings{wang2020truly,
  title={Truly proximal policy optimization},
  author={Wang, Yuhui and He, Hao and Tan, Xiaoyang},
  booktitle={UAI},
  year={2020},
}

@inproceedings{schulman2015high,
  title={High-dimensional continuous control using generalized advantage estimation},
  author={Schulman, John and Moritz, Philipp and Levine, Sergey and Jordan, Michael and Abbeel, Pieter},
  booktitle={ICLR},
  year={2016}
}

@inproceedings{brown2020language,
  title={Language models are few-shot learners},
  author={Brown, Tom and Mann, Benjamin and Ryder, Nick and Subbiah, Melanie and Kaplan, Jared D and Dhariwal, Prafulla and Neelakantan, Arvind and Shyam, Pranav and Sastry, Girish and Askell, Amanda and others},
  booktitle={NeurIPS},
  year={2020}
}

@article{yang2024qwen2,
  title={Qwen2. 5 technical report},
  author={Yang, An and Yang, Baosong and Zhang, Beichen and Hui, Binyuan and Zheng, Bo and Yu, Bowen and Li, Chengyuan and Liu, Dayiheng and Huang, Fei and Wei, Haoran and others},
  journal={arXiv:2412.15115},
  year={2024}
}

@article{grattafiori2024llama,
  title={The llama 3 herd of models},
  author={Grattafiori, Aaron and Dubey, Abhimanyu and Jauhri, Abhinav and Pandey, Abhinav and Kadian, Abhishek and Al-Dahle, Ahmad and Letman, Aiesha and Mathur, Akhil and Schelten, Alan and Vaughan, Alex and others},
  journal={arXiv:2407.21783},
  year={2024}
}

@article{wang2024qwen2,
  title={Qwen2-vl: Enhancing vision-language model's perception of the world at any resolution},
  author={Wang, Peng and Bai, Shuai and Tan, Sinan and Wang, Shijie and Fan, Zhihao and Bai, Jinze and Chen, Keqin and Liu, Xuejing and Wang, Jialin and Ge, Wenbin and others},
  journal={arXiv:2409.12191},
  year={2024}
}

@article{bai2025qwen2,
  title={Qwen2. 5-vl technical report},
  author={Bai, Shuai and Chen, Keqin and Liu, Xuejing and Wang, Jialin and Ge, Wenbin and Song, Sibo and Dang, Kai and Wang, Peng and Wang, Shijie and Tang, Jun and others},
  journal={arXiv:2502.13923},
  year={2025}
}

@article{zhu2025internvl3,
  title={InternVL3: Exploring Advanced Training and Test-Time Recipes for Open-Source Multimodal Models},
  author={Zhu, Jinguo and Wang, Weiyun and Chen, Zhe and Liu, Zhaoyang and Ye, Shenglong and Gu, Lixin and Duan, Yuchen and Tian, Hao and Su, Weijie and Shao, Jie and others},
  journal={arXiv:2504.10479},
  year={2025}
}

@article{li2023videochat,
  title={Videochat: Chat-centric video understanding},
  author={Li, KunChang and He, Yinan and Wang, Yi and Li, Yizhuo and Wang, Wenhai and Luo, Ping and Wang, Yali and Wang, Limin and Qiao, Yu},
  journal={arXiv preprint arXiv:2305.06355},
  year={2023}
}

@inproceedings{xu2023mplug,
  title={mplug-2: A modularized multi-modal foundation model across text, image and video},
  author={Xu, Haiyang and Ye, Qinghao and Yan, Ming and Shi, Yaya and Ye, Jiabo and Xu, Yuanhong and Li, Chenliang and Bi, Bin and Qian, Qi and Wang, Wei and others},
  booktitle={ICML},
  year={2023},
}

@article{li2024llava,
  title={Llava-onevision: Easy visual task transfer},
  author={Li, Bo and Zhang, Yuanhan and Guo, Dong and Zhang, Renrui and Li, Feng and Zhang, Hao and Zhang, Kaichen and Zhang, Peiyuan and Li, Yanwei and Liu, Ziwei and others},
  journal={TMLR},
  year={2025}
}

@article{cheng2024videollama,
  title={Videollama 2: Advancing spatial-temporal modeling and audio understanding in video-llms},
  author={Cheng, Zesen and Leng, Sicong and Zhang, Hang and Xin, Yifei and Li, Xin and Chen, Guanzheng and Zhu, Yongxin and Zhang, Wenqi and Luo, Ziyang and Zhao, Deli and others},
  journal={arXiv:2406.07476},
  year={2024}
}

@article{zhang2025videollama,
  title={VideoLLaMA 3: Frontier Multimodal Foundation Models for Image and Video Understanding},
  author={Zhang, Boqiang and Li, Kehan and Cheng, Zesen and Hu, Zhiqiang and Yuan, Yuqian and Chen, Guanzheng and Leng, Sicong and Jiang, Yuming and Zhang, Hang and Li, Xin and others},
  journal={arXiv:2501.13106},
  year={2025}
}

@article{zhang2024video,
  title={Video instruction tuning with synthetic data},
  author={Zhang, Yuanhan and Wu, Jinming and Li, Wei and Li, Bo and Ma, Zejun and Liu, Ziwei and Li, Chunyuan},
  journal={TMLR},
  year={2025}
}

@article{yu2025dapo,
  title={Dapo: An open-source llm reinforcement learning system at scale},
  author={Yu, Qiying and Zhang, Zheng and Zhu, Ruofei and Yuan, Yufeng and Zuo, Xiaochen and Yue, Yu and Dai, Weinan and Fan, Tiantian and Liu, Gaohong and Liu, Lingjun and others},
  journal={arXiv:2503.14476},
  year={2025}
}

@article{jaech2024openai,
  title={Openai o1 system card},
  author={Jaech, Aaron and Kalai, Adam and Lerer, Adam and Richardson, Adam and El-Kishky, Ahmed and Low, Aiden and Helyar, Alec and Madry, Aleksander and Beutel, Alex and Carney, Alex and others},
  journal={arXiv:2412.16720},
  year={2024}
}

@article{guo2025deepseek,
  title={Deepseek-r1: Incentivizing reasoning capability in llms via reinforcement learning},
  author={Guo, Daya and Yang, Dejian and Zhang, Haowei and Song, Junxiao and Zhang, Ruoyu and Xu, Runxin and Zhu, Qihao and Ma, Shirong and Wang, Peiyi and Bi, Xiao and others},
  journal={arXiv:2501.12948},
  year={2025}
}

@article{shao2024deepseekmath,
  title={Deepseekmath: Pushing the limits of mathematical reasoning in open language models},
  author={Shao, Zhihong and Wang, Peiyi and Zhu, Qihao and Xu, Runxin and Song, Junxiao and Bi, Xiao and Zhang, Haowei and Zhang, Mingchuan and Li, YK and Wu, Y and others},
  journal={arXiv:2402.03300},
  year={2024}
}

@article{team2025kimi,
  title={Kimi k1. 5: Scaling reinforcement learning with llms},
  author={Team, Kimi and Du, Angang and Gao, Bofei and Xing, Bowei and Jiang, Changjiu and Chen, Cheng and Li, Cheng and Xiao, Chenjun and Du, Chenzhuang and Liao, Chonghua and others},
  journal={arXiv:2501.12599},
  year={2025}
}

@inproceedings{liu2025visual,
  title={Visual-rft: Visual reinforcement fine-tuning},
  author={Liu, Ziyu and Sun, Zeyi and Zang, Yuhang and Dong, Xiaoyi and Cao, Yuhang and Duan, Haodong and Lin, Dahua and Wang, Jiaqi},
  booktitle={ICCV},
  year={2025}
}

@article{yang2025r1,
  title={R1-onevision: Advancing generalized multimodal reasoning through cross-modal formalization},
  author={Yang, Yi and He, Xiaoxuan and Pan, Hongkun and Jiang, Xiyan and Deng, Yan and Yang, Xingtao and Lu, Haoyu and Yin, Dacheng and Rao, Fengyun and Zhu, Minfeng and others},
  journal={arXiv:2503.10615},
  year={2025}
}

@article{zhan2025vision,
  title={Vision-R1: Evolving Human-Free Alignment in Large Vision-Language Models via Vision-Guided Reinforcement Learning},
  author={Zhan, Yufei and Zhu, Yousong and Zheng, Shurong and Zhao, Hongyin and Yang, Fan and Tang, Ming and Wang, Jinqiao},
  journal={arXiv:2503.18013},
  year={2025}
}

@article{huang2025vision,
  title={Vision-r1: Incentivizing reasoning capability in multimodal large language models},
  author={Huang, Wenxuan and Jia, Bohan and Zhai, Zijie and Cao, Shaosheng and Ye, Zheyu and Zhao, Fei and Xu, Zhe and Hu, Yao and Lin, Shaohui},
  journal={arXiv:2503.06749},
  year={2025}
}

@inproceedings{huang2025boosting,
  title={Boosting MLLM Reasoning with Text-Debiased Hint-GRPO},
  author={Huang, Qihan and Chan, Long and Liu, Jinlong and He, Wanggui and Jiang, Hao and Song, Mingli and Chen, Jingyuan and Yao, Chang and Song, Jie},
  booktitle={ICCV},
  year={2025}
}

@article{shen2025vlm,
  title={Vlm-r1: A stable and generalizable r1-style large vision-language model},
  author={Shen, Haozhan and Liu, Peng and Li, Jingcheng and Fang, Chunxin and Ma, Yibo and Liao, Jiajia and Shen, Qiaoli and Zhang, Zilun and Zhao, Kangjia and Zhang, Qianqian and others},
  journal={arXiv:2504.07615},
  year={2025}
}

@inproceedings{liu2025noisyrollout,
  title={NoisyRollout: Reinforcing Visual Reasoning with Data Augmentation},
  author={Liu, Xiangyan and Ni, Jinjie and Wu, Zijian and Du, Chao and Dou, Longxu and Wang, Haonan and Pang, Tianyu and Shieh, Michael Qizhe},
  booktitle={NeurIPS},
  year={2025}
}

@inproceedings{tan2025reason,
  title={Reason-rft: Reinforcement fine-tuning for visual reasoning},
  author={Tan, Huajie and Ji, Yuheng and Hao, Xiaoshuai and Lin, Minglan and Wang, Pengwei and Wang, Zhongyuan and Zhang, Shanghang},
  booktitle={NeurIPS},
  year={2025}
}

@inproceedings{feng2025video,
  title={Video-r1: Reinforcing video reasoning in mllms},
  author={Feng, Kaituo and Gong, Kaixiong and Li, Bohao and Guo, Zonghao and Wang, Yibing and Peng, Tianshuo and Wang, Benyou and Yue, Xiangyu},
  booktitle={NeurIPS},
  year={2025}
}

@inproceedings{li2025videochat,
  title={VideoChat-R1: Enhancing Spatio-Temporal Perception via Reinforcement Fine-Tuning},
  author={Li, Xinhao and Yan, Ziang and Meng, Desen and Dong, Lu and Zeng, Xiangyu and He, Yinan and Wang, Yali and Qiao, Yu and Wang, Yi and Wang, Limin},
  booktitle={NeurIPS},
  year={2025}
}

@article{chen2025exploring,
  title={Exploring the Effect of Reinforcement Learning on Video Understanding: Insights from SEED-Bench-R1},
  author={Chen, Yi and Ge, Yuying and Wang, Rui and Ge, Yixiao and Qiu, Lu and Shan, Ying and Liu, Xihui},
  journal={arXiv:2503.24376},
  year={2025}
}

@article{wang2025timezero,
  title={TimeZero: Temporal Video Grounding with Reasoning-Guided LVLM},
  author={Wang, Ye and Xu, Boshen and Yue, Zihao and Xiao, Zihan and Wang, Ziheng and Zhang, Liang and Yang, Dingyi and Wang, Wenxuan and Jin, Qin},
  journal={arXiv:2503.13377},
  year={2025}
}

@article{wu2025st,
  title={ST-Think: How multimodal large language models reason about 4d worlds from ego-centric videos},
  author={Wu, Peiran and Liu, Yunze and Liu, Miao and Shen, Junxiao},
  journal={arXiv:2503.12542},
  year={2025}
}

@inproceedings{wu2024longvideobench,
  title={Longvideobench: A benchmark for long-context interleaved video-language understanding},
  author={Wu, Haoning and Li, Dongxu and Chen, Bei and Li, Junnan},
  booktitle={NeurIPS},
  year={2024}
}

@inproceedings{fu2024video,
  title={Video-mme: The first-ever comprehensive evaluation benchmark of multi-modal llms in video analysis},
  author={Fu, Chaoyou and Dai, Yuhan and Luo, Yongdong and Li, Lei and Ren, Shuhuai and Zhang, Renrui and Wang, Zihan and Zhou, Chenyu and Shen, Yunhang and Zhang, Mengdan and others},
  booktitle={CVPR},
  year={2025}
}

@inproceedings{li2024mvbench,
  title={Mvbench: A comprehensive multi-modal video understanding benchmark},
  author={Li, Kunchang and Wang, Yali and He, Yinan and Li, Yizhuo and Wang, Yi and Liu, Yi and Wang, Zun and Xu, Jilan and Chen, Guo and Luo, Ping and others},
  booktitle={CVPR},
  year={2024}
}

@inproceedings{gao2017tall,
  title={Tall: Temporal activity localization via language query},
  author={Gao, Jiyang and Sun, Chen and Yang, Zhenheng and Nevatia, Ram},
  booktitle={ICCV},
  year={2017}
}

@inproceedings{krishna2017dense,
  title={Dense-captioning events in videos},
  author={Krishna, Ranjay and Hata, Kenji and Ren, Frederic and Fei-Fei, Li and Carlos Niebles, Juan},
  booktitle={ICCV},
  year={2017}
}

@inproceedings{xiao2024can,
  title={Can i trust your answer? visually grounded video question answering},
  author={Xiao, Junbin and Yao, Angela and Li, Yicong and Chua, Tat-Seng},
  booktitle={CVPR},
  year={2024}
}

@article{damen2022rescaling,
  title={Rescaling egocentric vision: Collection, pipeline and challenges for epic-kitchens-100},
  author={Damen, Dima and Doughty, Hazel and Farinella, Giovanni Maria and Furnari, Antonino and Kazakos, Evangelos and Ma, Jian and Moltisanti, Davide and Munro, Jonathan and Perrett, Toby and Price, Will and others},
  journal={IJCV},
  pages={1--23},
  year={2022},
}

@inproceedings{grauman2022ego4d,
  title={Ego4d: Around the world in 3,000 hours of egocentric video},
  author={Grauman, Kristen and Westbury, Andrew and Byrne, Eugene and Chavis, Zachary and Furnari, Antonino and Girdhar, Rohit and Hamburger, Jackson and Jiang, Hao and Liu, Miao and Liu, Xingyu and others},
  booktitle={CVPR},
  year={2022}
}

@article{chen2023egoplan,
  title={Egoplan-bench: Benchmarking multimodal large language models for human-level planning},
  author={Chen, Yi and Ge, Yuying and Ge, Yixiao and Ding, Mingyu and Li, Bohao and Wang, Rui and Xu, Ruifeng and Shan, Ying and Liu, Xihui},
  journal={arXiv:2312.06722},
  year={2023}
}

@article{qiu2024egoplan,
  title={Egoplan-bench2: A benchmark for multimodal large language model planning in real-world scenarios},
  author={Qiu, Lu and Chen, Yi and Ge, Yuying and Ge, Yixiao and Shan, Ying and Liu, Xihui},
  journal={arXiv:2412.04447},
  year={2024}
}

@inproceedings{huang2024n+,
  title={The n+ implementation details of rlhf with ppo: A case study on tl; dr summarization},
  author={Huang, Shengyi and Noukhovitch, Michael and Hosseini, Arian and Rasul, Kashif and Wang, Weixun and Tunstall, Lewis},
  booktitle={COLM},
  year={2024}
}

@inproceedings{ouyang2022training,
  title={Training language models to follow instructions with human feedback},
  author={Ouyang, Long and Wu, Jeffrey and Jiang, Xu and Almeida, Diogo and Wainwright, Carroll and Mishkin, Pamela and Zhang, Chong and Agarwal, Sandhini and Slama, Katarina and Ray, Alex and others},
  booktitle={NeurIPS},
  year={2022}
}

@inproceedings{wang2025vl,
  title={VL-Rethinker: Incentivizing Self-Reflection of Vision-Language Models with Reinforcement Learning},
  author={Wang, Haozhe and Qu, Chao and Huang, Zuming and Chu, Wei and Lin, Fangzhen and Chen, Wenhu},
  booktitle={NeurIPS},
  year={2025}
}

@article{touvron2023llama,
  title={Llama 2: Open foundation and fine-tuned chat models},
  author={Touvron, Hugo and Martin, Louis and Stone, Kevin and Albert, Peter and Almahairi, Amjad and Babaei, Yasmine and Bashlykov, Nikolay and Batra, Soumya and Bhargava, Prajjwal and Bhosale, Shruti and others},
  journal={arXiv:2307.09288},
  year={2023}
}

@article{achiam2023gpt,
  title={Gpt-4 technical report},
  author={Achiam, Josh and Adler, Steven and Agarwal, Sandhini and Ahmad, Lama and Akkaya, Ilge and Aleman, Florencia Leoni and Almeida, Diogo and Altenschmidt, Janko and Altman, Sam and Anadkat, Shyamal and others},
  journal={arXiv:2303.08774},
  year={2023}
}

@article{team2023gemini,
  title={Gemini: a family of highly capable multimodal models},
  author={Team, Gemini and Anil, Rohan and Borgeaud, Sebastian and Wu, Yonghui and Alayrac, Jean-Baptiste and Yu, Jiahui and Soricut, Radu and Schalkwyk, Johan and Dai, Andrew M and Hauth, Anja and others},
  journal={arXiv:2312.11805},
  year={2023}
}

@inproceedings{radford2021learning,
  title={Learning transferable visual models from natural language supervision},
  author={Radford, Alec and Kim, Jong Wook and Hallacy, Chris and Ramesh, Aditya and Goh, Gabriel and Agarwal, Sandhini and Sastry, Girish and Askell, Amanda and Mishkin, Pamela and Clark, Jack and others},
  booktitle={ICML},
  year={2021}
}

@inproceedings{wang2024internvideo2,
  title={Internvideo2: Scaling foundation models for multimodal video understanding},
  author={Wang, Yi and Li, Kunchang and Li, Xinhao and Yu, Jiashuo and He, Yinan and Chen, Guo and Pei, Baoqi and Zheng, Rongkun and Wang, Zun and Shi, Yansong and others},
  booktitle={ECCV},
  year={2024},
}

@inproceedings{yu2023self,
  title={Self-chained image-language model for video localization and question answering},
  author={Yu, Shoubin and Cho, Jaemin and Yadav, Prateek and Bansal, Mohit},
  booktitle={NeurIPS},
  year={2023}
}

@inproceedings{peters2007reinforcement,
  title={Reinforcement learning by reward-weighted regression for operational space control},
  author={Peters, Jan and Schaal, Stefan},
  booktitle={ICML},
  year={2007}
}

@inproceedings{peters2010relative,
  title={Relative entropy policy search},
  author={Peters, Jan and Mulling, Katharina and Altun, Yasemin},
  booktitle={AAAI},
  year={2010}
}

@inproceedings{zhang2024direct,
  title={Direct preference optimization of video large multimodal models from language model reward},
  author={Zhang, Ruohong and Gui, Liangke and Sun, Zhiqing and Feng, Yihao and Xu, Keyang and Zhang, Yuanhan and Fu, Di and Li, Chunyuan and Hauptmann, Alexander and Bisk, Yonatan and others},
  booktitle={NAACL},
  year={2025}
}

@inproceedings{ahn2025isr,
  title={ISR-DPO: Aligning Large Multimodal Models for Videos by Iterative Self-Retrospective DPO},
  author={Ahn, Daechul and Choi, Yura and Kim, San and Yu, Youngjae and Kang, Dongyeop and Choi, Jonghyun},
  booktitle={AAAI},
  year={2025}
}

@inproceedings{hsu2020revisiting,
  title={Revisiting design choices in proximal policy optimization},
  author={Hsu, Chloe Ching-Yun and Mendler-D{\"u}nner, Celestine and Hardt, Moritz},
  booktitle={ICLR},
  year={2020}
}

@inproceedings{li2022invariant,
  title={Invariant grounding for video question answering},
  author={Li, Yicong and Wang, Xiang and Xiao, Junbin and Ji, Wei and Chua, Tat-Seng},
  booktitle={CVPR},
  year={2022}
}

@inproceedings{yang2022zero,
  title={Zero-shot video question answering via frozen bidirectional language models},
  author={Yang, Antoine and Miech, Antoine and Sivic, Josef and Laptev, Ivan and Schmid, Cordelia},
  booktitle={NeurIPS},
  year={2022}
}

@inproceedings{li2024llama,
  title={Llama-vid: An image is worth 2 tokens in large language models},
  author={Li, Yanwei and Wang, Chengyao and Jia, Jiaya},
  booktitle={ECCV},
  year={2024},
}

@article{zhang2024long,
  title={Long context transfer from language to vision},
  author={Zhang, Peiyuan and Zhang, Kaichen and Li, Bo and Zeng, Guangtao and Yang, Jingkang and Zhang, Yuanhan and Wang, Ziyue and Tan, Haoran and Li, Chunyuan and Liu, Ziwei},
  journal={TMLR},
  year={2025}
}

@inproceedings{lin2024vila,
  title={Vila: On pre-training for visual language models},
  author={Lin, Ji and Yin, Hongxu and Ping, Wei and Molchanov, Pavlo and Shoeybi, Mohammad and Han, Song},
  booktitle={CVPR},
  year={2024}
}

@inproceedings{yu2025unhackable,
  title={Unhackable temporal rewarding for scalable video mllms},
  author={Yu, En and Lin, Kangheng and Zhao, Liang and Wei, Yana and Zhu, Zining and Wei, Haoran and Sun, Jianjian and Ge, Zheng and Zhang, Xiangyu and Wang, Jingyu and others},
  booktitle={ICLR},
  year={2025}
}

@article{liu2024kangaroo,
  title={Kangaroo: A powerful video-language model supporting long-context video input},
  author={Liu, Jiajun and Wang, Yibing and Ma, Hanghang and Wu, Xiaoping and Ma, Xiaoqi and Wei, Xiaoming and Jiao, Jianbin and Wu, Enhua and Hu, Jie},
  journal={arXiv:2408.15542},
  year={2024}
}

@inproceedings{yang2024thinking,
  title={Thinking in space: How multimodal large language models see, remember, and recall spaces},
  author={Yang, Jihan and Yang, Shusheng and Gupta, Anjali W and Han, Rilyn and Fei-Fei, Li and Xie, Saining},
  booktitle={CVPR},
  year={2025}
}

@article{hu2025video,
  title={Video-MMMU: Evaluating Knowledge Acquisition from Multi-Discipline Professional Videos},
  author={Hu, Kairui and Wu, Penghao and Pu, Fanyi and Xiao, Wang and Zhang, Yuanhan and Yue, Xiang and Li, Bo and Liu, Ziwei},
  journal={arXiv:2501.13826},
  year={2025}
}

@inproceedings{zhao2025mmvu,
  title={MMVU: Measuring Expert-Level Multi-Discipline Video Understanding},
  author={Zhao, Yilun and Xie, Lujing and Zhang, Haowei and Gan, Guo and Long, Yitao and Hu, Zhiyuan and Hu, Tongyan and Chen, Weiyuan and Li, Chuhan and Song, Junyang and others},
  booktitle={CVPR},
  year={2025}
}

@inproceedings{liu2024tempcompass,
  title={Tempcompass: Do video llms really understand videos?},
  author={Liu, Yuanxin and Li, Shicheng and Liu, Yi and Wang, Yuxiang and Ren, Shuhuai and Li, Lei and Chen, Sishuo and Sun, Xu and Hou, Lu},
  booktitle={ACL-Findings},
  year={2024}
}

@inproceedings{paszke2019pytorch,
  title={Pytorch: An imperative style, high-performance deep learning library},
  author={Paszke, Adam and Gross, Sam and Massa, Francisco and Lerer, Adam and Bradbury, James and Chanan, Gregory and Killeen, Trevor and Lin, Zeming and Gimelshein, Natalia and Antiga, Luca and others},
  booktitle={NeurIPS},
  year={2019}
}

@inproceedings{wolf2019huggingface,
  title={Huggingface's transformers: State-of-the-art natural language processing},
  author={Wolf, Thomas and Debut, Lysandre and Sanh, Victor and Chaumond, Julien and Delangue, Clement and Moi, Anthony and Cistac, Pierric and Rault, Tim and Louf, R{\'e}mi and Funtowicz, Morgan and others},
  booktitle={EMNLP demo},
  year={2020}
}

@misc{vonwerra2022trl,
  author = {Leandro von Werra and Younes Belkada and Lewis Tunstall and Edward Beeching and Tristan Thrush and Nathan Lambert and Shengyi Huang and Kashif Rasul and Quentin Gallouédec},
  title = {TRL: Transformer Reinforcement Learning},
  year = {2020},
  publisher = {GitHub},
  journal = {GitHub repository},
  howpublished = {\url{https://github.com/huggingface/trl}}
}

@inproceedings{kwon2023efficient,
  title={Efficient Memory Management for Large Language Model Serving with PagedAttention},
  author={Woosuk Kwon and Zhuohan Li and Siyuan Zhuang and Ying Sheng and Lianmin Zheng and Cody Hao Yu and Joseph E. Gonzalez and Hao Zhang and Ion Stoica},
  booktitle={SOSP},
  year={2023}
}

@inproceedings{kreutzer2017bandit,
  title={Bandit structured prediction for neural sequence-to-sequence learning},
  author={Kreutzer, Julia and Sokolov, Artem and Riezler, Stefan},
  booktitle={ACL},
  year={2017}
}

@inproceedings{nguyen2017reinforcement,
  title={Reinforcement learning for bandit neural machine translation with simulated human feedback},
  author={Nguyen, Khanh and Daum{\'e} III, Hal and Boyd-Graber, Jordan},
  booktitle={EMNLP},
  year={2017}
}

@article{williams1992simple,
  title={Simple statistical gradient-following algorithms for connectionist reinforcement learning},
  author={Williams, Ronald J},
  journal={Machine learning},
  volume={8},
  year={1992},
}

@inproceedings{kool2019buy,
  title={Buy 4 reinforce samples, get a baseline for free!},
  author={Kool, Wouter and van Hoof, Herke and Welling, Max},
  booktitle={ICLRW},
  year={2019}
}

@inproceedings{park2024llamo,
  title={Llamo: Large language model-based molecular graph assistant},
  author={Park, Jinyoung and Bae, Minseong and Ko, Dohwan and Kim, Hyunwoo J},
  booktitle={NeurIPS},
  year={2024}
}

@article{park2023graph,
  title={Graph elicitation for guiding multi-step reasoning in large language models},
  author={Park, Jinyoung and Patel, Ameen and Khan, Omar Zia and Kim, Hyunwoo J and Kim, Joo-Kyung},
  journal={arXiv:2311.09762},
  year={2023}
}

@inproceedings{park2024generative,
  title={Generative subgraph retrieval for knowledge graph-grounded dialog generation},
  author={Park, Jinyoung and Joo, Minseok and Kim, Joo-Kyung and Kim, Hyunwoo J},
  booktitle={EMNLP},
  year={2024}
}

@inproceedings{lee2025vidchain,
  title={VidChain: Chain-of-Tasks with Metric-based Direct Preference Optimization for Dense Video Captioning},
  author={Lee, Ji Soo and Kim, Jongha and Na, Jeehye and Park, Jinyoung and Kim, Hyunwoo J},
  booktitle={AAAI},
  year={2025}
}

@inproceedings{ko2023large,
  title={Large language models are temporal and causal reasoners for video question answering},
  author={Ko, Dohwan and Lee, Ji Soo and Kang, Wooyoung and Roh, Byungseok and Kim, Hyunwoo J},
  booktitle={EMNLP},
  year={2023}
}

@inproceedings{shen2024longvu,
  title={Longvu: Spatiotemporal adaptive compression for long video-language understanding},
  author={Shen, Xiaoqian and Xiong, Yunyang and Zhao, Changsheng and Wu, Lemeng and Chen, Jun and Zhu, Chenchen and Liu, Zechun and Xiao, Fanyi and Varadarajan, Balakrishnan and Bordes, Florian and others},
  booktitle={ICML},
  year={2025}
}

@inproceedings{lee2025captioning,
  title={Captioning for Text-Video Retrieval via Dual-Group Direct Preference Optimization},
  author={Lee, Ji Soo and Ko, Byungoh and Cho, Jaewon and Lee, Howoong and Byun, Jaewoon and Kim, Hyunwoo J},
  booktitle={EMNLP-Findings},
  year={2025}
}

@inproceedings{luo2023wizardmath,
  title={Wizardmath: Empowering mathematical reasoning for large language models via reinforced evol-instruct},
  author={Luo, Haipeng and Sun, Qingfeng and Xu, Can and Zhao, Pu and Lou, Jianguang and Tao, Chongyang and Geng, Xiubo and Lin, Qingwei and Chen, Shifeng and Zhang, Dongmei},
  booktitle={ICLR},
  year={2025}
}

@article{bai2025intern,
  title={Intern-s1: A scientific multimodal foundation model},
  author={Bai, Lei and Cai, Zhongrui and Cao, Yuhang and Cao, Maosong and Cao, Weihan and Chen, Chiyu and Chen, Haojiong and Chen, Kai and Chen, Pengcheng and Chen, Ying and others},
  journal={arXiv:2508.15763},
  year={2025}
}

@article{zheng2025large,
  title={Large language models for scientific discovery in molecular property prediction},
  author={Zheng, Yizhen and Koh, Huan Yee and Ju, Jiaxin and Nguyen, Anh TN and May, Lauren T and Webb, Geoffrey I and Pan, Shirui},
  journal={Nat. Mach. Intell.},
  pages={1--11},
  year={2025},
}

@inproceedings{jin2025search,
  title={Search-r1: Training llms to reason and leverage search engines with reinforcement learning},
  author={Jin, Bowen and Zeng, Hansi and Yue, Zhenrui and Yoon, Jinsung and Arik, Sercan and Wang, Dong and Zamani, Hamed and Han, Jiawei},
  booktitle={COLM},
  year={2025}
}

@article{luo2025graph,
  title={Graph-r1: Towards agentic graphrag framework via end-to-end reinforcement learning},
  author={Luo, Haoran and Chen, Guanting and Lin, Qika and Guo, Yikai and Xu, Fangzhi and Kuang, Zemin and Song, Meina and Wu, Xiaobao and Zhu, Yifan and Tuan, Luu Anh and others},
  journal={arXiv:2507.21892},
  year={2025}
}
}





\newpage
\section*{NeurIPS Paper Checklist}

\begin{enumerate}

  \item {\bf Claims}
  \item[] Question: Do the main claims made in the abstract and introduction accurately reflect the paper's contributions and scope?
  \item[] Answer: \answerYes{} 
  \item[] Justification: In Abstract and Introduction.
  \item[] Guidelines:
    \begin{itemize}
      \item The answer NA means that the abstract and introduction do not include the claims made in the paper.
      \item The abstract and/or introduction should clearly state the claims made, including the contributions made in the paper and important assumptions and limitations. A No or NA answer to this question will not be perceived well by the reviewers.
      \item The claims made should match theoretical and experimental results, and reflect how much the results can be expected to generalize to other settings.
      \item It is fine to include aspirational goals as motivation as long as it is clear that these goals are not attained by the paper.
    \end{itemize}

  \item {\bf Limitations}
  \item[] Question: Does the paper discuss the limitations of the work performed by the authors?
  \item[] Answer: \answerYes{} 
  \item[] Justification: In the appendix.
  \item[] Guidelines:
    \begin{itemize}
      \item The answer NA means that the paper has no limitation while the answer No means that the paper has limitations, but those are not discussed in the paper.
      \item The authors are encouraged to create a separate "Limitations" section in their paper.
      \item The paper should point out any strong assumptions and how robust the results are to violations of these assumptions (e.g., independence assumptions, noiseless settings, model well-specification, asymptotic approximations only holding locally). The authors should reflect on how these assumptions might be violated in practice and what the implications would be.
      \item The authors should reflect on the scope of the claims made, e.g., if the approach was only tested on a few datasets or with a few runs. In general, empirical results often depend on implicit assumptions, which should be articulated.
      \item The authors should reflect on the factors that influence the performance of the approach. For example, a facial recognition algorithm may perform poorly when image resolution is low or images are taken in low lighting. Or a speech-to-text system might not be used reliably to provide closed captions for online lectures because it fails to handle technical jargon.
      \item The authors should discuss the computational efficiency of the proposed algorithms and how they scale with dataset size.
      \item If applicable, the authors should discuss possible limitations of their approach to address problems of privacy and fairness.
      \item While the authors might fear that complete honesty about limitations might be used by reviewers as grounds for rejection, a worse outcome might be that reviewers discover limitations that aren't acknowledged in the paper. The authors should use their best judgment and recognize that individual actions in favor of transparency play an important role in developing norms that preserve the integrity of the community. Reviewers will be specifically instructed to not penalize honesty concerning limitations.
    \end{itemize}

  \item {\bf Theory assumptions and proofs}
  \item[] Question: For each theoretical result, does the paper provide the full set of assumptions and a complete (and correct) proof?
  \item[] Answer: \answerYes{} 
  \item[] Justification: We provide the assumptions and derivation in the appendix.
  \item[] Guidelines:
    \begin{itemize}
      \item The answer NA means that the paper does not include theoretical results.
      \item All the theorems, formulas, and proofs in the paper should be numbered and cross-referenced.
      \item All assumptions should be clearly stated or referenced in the statement of any theorems.
      \item The proofs can either appear in the main paper or the supplemental material, but if they appear in the supplemental material, the authors are encouraged to provide a short proof sketch to provide intuition.
      \item Inversely, any informal proof provided in the core of the paper should be complemented by formal proofs provided in appendix or supplemental material.
      \item Theorems and Lemmas that the proof relies upon should be properly referenced.
    \end{itemize}

  \item {\bf Experimental result reproducibility}
  \item[] Question: Does the paper fully disclose all the information needed to reproduce the main experimental results of the paper to the extent that it affects the main claims and/or conclusions of the paper (regardless of whether the code and data are provided or not)?
  \item[] Answer: \answerYes{} 
  \item[] Justification: In Experiments and appendix.
  \item[] Guidelines:
    \begin{itemize}
      \item The answer NA means that the paper does not include experiments.
      \item If the paper includes experiments, a No answer to this question will not be perceived well by the reviewers: Making the paper reproducible is important, regardless of whether the code and data are provided or not.
      \item If the contribution is a dataset and/or model, the authors should describe the steps taken to make their results reproducible or verifiable.
      \item Depending on the contribution, reproducibility can be accomplished in various ways. For example, if the contribution is a novel architecture, describing the architecture fully might suffice, or if the contribution is a specific model and empirical evaluation, it may be necessary to either make it possible for others to replicate the model with the same dataset, or provide access to the model. In general. releasing code and data is often one good way to accomplish this, but reproducibility can also be provided via detailed instructions for how to replicate the results, access to a hosted model (e.g., in the case of a large language model), releasing of a model checkpoint, or other means that are appropriate to the research performed.
      \item While NeurIPS does not require releasing code, the conference does require all submissions to provide some reasonable avenue for reproducibility, which may depend on the nature of the contribution. For example
        \begin{enumerate}
          \item If the contribution is primarily a new algorithm, the paper should make it clear how to reproduce that algorithm.
          \item If the contribution is primarily a new model architecture, the paper should describe the architecture clearly and fully.
          \item If the contribution is a new model (e.g., a large language model), then there should either be a way to access this model for reproducing the results or a way to reproduce the model (e.g., with an open-source dataset or instructions for how to construct the dataset).
          \item We recognize that reproducibility may be tricky in some cases, in which case authors are welcome to describe the particular way they provide for reproducibility. In the case of closed-source models, it may be that access to the model is limited in some way (e.g., to registered users), but it should be possible for other researchers to have some path to reproducing or verifying the results.
        \end{enumerate}
    \end{itemize}

  \item {\bf Open access to data and code}
  \item[] Question: Does the paper provide open access to the data and code, with sufficient instructions to faithfully reproduce the main experimental results, as described in supplemental material?
  \item[] Answer: \answerYes{} 
  \item[] Justification: In Abstract.
  \item[] Guidelines:
    \begin{itemize}
      \item The answer NA means that paper does not include experiments requiring code.
      \item Please see the NeurIPS code and data submission guidelines (\url{https://nips.cc/public/guides/CodeSubmissionPolicy}) for more details.
      \item While we encourage the release of code and data, we understand that this might not be possible, so “No” is an acceptable answer. Papers cannot be rejected simply for not including code, unless this is central to the contribution (e.g., for a new open-source benchmark).
      \item The instructions should contain the exact command and environment needed to run to reproduce the results. See the NeurIPS code and data submission guidelines (\url{https://nips.cc/public/guides/CodeSubmissionPolicy}) for more details.
      \item The authors should provide instructions on data access and preparation, including how to access the raw data, preprocessed data, intermediate data, and generated data, etc.
      \item The authors should provide scripts to reproduce all experimental results for the new proposed method and baselines. If only a subset of experiments are reproducible, they should state which ones are omitted from the script and why.
      \item At submission time, to preserve anonymity, the authors should release anonymized versions (if applicable).
      \item Providing as much information as possible in supplemental material (appended to the paper) is recommended, but including URLs to data and code is permitted.
    \end{itemize}

  \item {\bf Experimental setting/details}
  \item[] Question: Does the paper specify all the training and test details (e.g., data splits, hyperparameters, how they were chosen, type of optimizer, etc.) necessary to understand the results?
  \item[] Answer: \answerYes{} 
  \item[] Justification: In Experiment and the appendix.
  \item[] Guidelines:
    \begin{itemize}
      \item The answer NA means that the paper does not include experiments.
      \item The experimental setting should be presented in the core of the paper to a level of detail that is necessary to appreciate the results and make sense of them.
      \item The full details can be provided either with the code, in appendix, or as supplemental material.
    \end{itemize}

  \item {\bf Experiment statistical significance}
  \item[] Question: Does the paper report error bars suitably and correctly defined or other appropriate information about the statistical significance of the experiments?
  \item[] Answer: \answerNA{} 
  \item[] Justification: We conduct the experiment in the single run following other works.
  \item[] Guidelines:
    \begin{itemize}
      \item The answer NA means that the paper does not include experiments.
      \item The authors should answer "Yes" if the results are accompanied by error bars, confidence intervals, or statistical significance tests, at least for the experiments that support the main claims of the paper.
      \item The factors of variability that the error bars are capturing should be clearly stated (for example, train/test split, initialization, random drawing of some parameter, or overall run with given experimental conditions).
      \item The method for calculating the error bars should be explained (closed form formula, call to a library function, bootstrap, etc.)
      \item The assumptions made should be given (e.g., Normally distributed errors).
      \item It should be clear whether the error bar is the standard deviation or the standard error of the mean.
      \item It is OK to report 1-sigma error bars, but one should state it. The authors should preferably report a 2-sigma error bar than state that they have a 96\% CI, if the hypothesis of Normality of errors is not verified.
      \item For asymmetric distributions, the authors should be careful not to show in tables or figures symmetric error bars that would yield results that are out of range (e.g. negative error rates).
      \item If error bars are reported in tables or plots, The authors should explain in the text how they were calculated and reference the corresponding figures or tables in the text.
    \end{itemize}

  \item {\bf Experiments compute resources}
  \item[] Question: For each experiment, does the paper provide sufficient information on the computer resources (type of compute workers, memory, time of execution) needed to reproduce the experiments?
  \item[] Answer: \answerYes{} 
  \item[] Justification: In the appendix.
  \item[] Guidelines:
    \begin{itemize}
      \item The answer NA means that the paper does not include experiments.
      \item The paper should indicate the type of compute workers CPU or GPU, internal cluster, or cloud provider, including relevant memory and storage.
      \item The paper should provide the amount of compute required for each of the individual experimental runs as well as estimate the total compute.
      \item The paper should disclose whether the full research project required more compute than the experiments reported in the paper (e.g., preliminary or failed experiments that didn't make it into the paper).
    \end{itemize}

  \item {\bf Code of ethics}
  \item[] Question: Does the research conducted in the paper conform, in every respect, with the NeurIPS Code of Ethics \url{https://neurips.cc/public/EthicsGuidelines}?
  \item[] Answer: \answerYes{} 
  \item[] Justification: We follow NeurIPS Code of Ethics.
  \item[] Guidelines:
    \begin{itemize}
      \item The answer NA means that the authors have not reviewed the NeurIPS Code of Ethics.
      \item If the authors answer No, they should explain the special circumstances that require a deviation from the Code of Ethics.
      \item The authors should make sure to preserve anonymity (e.g., if there is a special consideration due to laws or regulations in their jurisdiction).
    \end{itemize}

  \item {\bf Broader impacts}
  \item[] Question: Does the paper discuss both potential positive societal impacts and negative societal impacts of the work performed?
  \item[] Answer: \answerYes{} 
  \item[] Justification: In the appendix.
  \item[] Guidelines:
    \begin{itemize}
      \item The answer NA means that there is no societal impact of the work performed.
      \item If the authors answer NA or No, they should explain why their work has no societal impact or why the paper does not address societal impact.
      \item Examples of negative societal impacts include potential malicious or unintended uses (e.g., disinformation, generating fake profiles, surveillance), fairness considerations (e.g., deployment of technologies that could make decisions that unfairly impact specific groups), privacy considerations, and security considerations.
      \item The conference expects that many papers will be foundational research and not tied to particular applications, let alone deployments. However, if there is a direct path to any negative applications, the authors should point it out. For example, it is legitimate to point out that an improvement in the quality of generative models could be used to generate deepfakes for disinformation. On the other hand, it is not needed to point out that a generic algorithm for optimizing neural networks could enable people to train models that generate Deepfakes faster.
      \item The authors should consider possible harms that could arise when the technology is being used as intended and functioning correctly, harms that could arise when the technology is being used as intended but gives incorrect results, and harms following from (intentional or unintentional) misuse of the technology.
      \item If there are negative societal impacts, the authors could also discuss possible mitigation strategies (e.g., gated release of models, providing defenses in addition to attacks, mechanisms for monitoring misuse, mechanisms to monitor how a system learns from feedback over time, improving the efficiency and accessibility of ML).
    \end{itemize}

  \item {\bf Safeguards}
  \item[] Question: Does the paper describe safeguards that have been put in place for responsible release of data or models that have a high risk for misuse (e.g., pretrained language models, image generators, or scraped datasets)?
  \item[] Answer: \answerNA{} 
  \item[] Justification: We do not release any datasets.
  \item[] Guidelines:
    \begin{itemize}
      \item The answer NA means that the paper poses no such risks.
      \item Released models that have a high risk for misuse or dual-use should be released with necessary safeguards to allow for controlled use of the model, for example by requiring that users adhere to usage guidelines or restrictions to access the model or implementing safety filters.
      \item Datasets that have been scraped from the Internet could pose safety risks. The authors should describe how they avoided releasing unsafe images.
      \item We recognize that providing effective safeguards is challenging, and many papers do not require this, but we encourage authors to take this into account and make a best faith effort.
    \end{itemize}

  \item {\bf Licenses for existing assets}
  \item[] Question: Are the creators or original owners of assets (e.g., code, data, models), used in the paper, properly credited and are the license and terms of use explicitly mentioned and properly respected?
  \item[] Answer: \answerYes{} 
  \item[] Justification: In the appendix.
  \item[] Guidelines:
    \begin{itemize}
      \item The answer NA means that the paper does not use existing assets.
      \item The authors should cite the original paper that produced the code package or dataset.
      \item The authors should state which version of the asset is used and, if possible, include a URL.
      \item The name of the license (e.g., CC-BY 4.0) should be included for each asset.
      \item For scraped data from a particular source (e.g., website), the copyright and terms of service of that source should be provided.
      \item If assets are released, the license, copyright information, and terms of use in the package should be provided. For popular datasets, \url{paperswithcode.com/datasets} has curated licenses for some datasets. Their licensing guide can help determine the license of a dataset.
      \item For existing datasets that are re-packaged, both the original license and the license of the derived asset (if it has changed) should be provided.
      \item If this information is not available online, the authors are encouraged to reach out to the asset's creators.
    \end{itemize}

  \item {\bf New assets}
  \item[] Question: Are new assets introduced in the paper well documented and is the documentation provided alongside the assets?
  \item[] Answer: \answerYes{} 
  \item[] Justification: In the appendix.
  \item[] Guidelines:
    \begin{itemize}
      \item The answer NA means that the paper does not release new assets.
      \item Researchers should communicate the details of the dataset/code/model as part of their submissions via structured templates. This includes details about training, license, limitations, etc.
      \item The paper should discuss whether and how consent was obtained from people whose asset is used.
      \item At submission time, remember to anonymize your assets (if applicable). You can either create an anonymized URL or include an anonymized zip file.
    \end{itemize}

  \item {\bf Crowdsourcing and research with human subjects}
  \item[] Question: For crowdsourcing experiments and research with human subjects, does the paper include the full text of instructions given to participants and screenshots, if applicable, as well as details about compensation (if any)?
  \item[] Answer: \answerNA{} 
  \item[] Justification: The paper does not involve crowdsourcing nor research with human subjects.
  \item[] Guidelines:
    \begin{itemize}
      \item The answer NA means that the paper does not involve crowdsourcing nor research with human subjects.
      \item Including this information in the supplemental material is fine, but if the main contribution of the paper involves human subjects, then as much detail as possible should be included in the main paper.
      \item According to the NeurIPS Code of Ethics, workers involved in data collection, curation, or other labor should be paid at least the minimum wage in the country of the data collector.
    \end{itemize}

  \item {\bf Institutional review board (IRB) approvals or equivalent for research with human subjects}
  \item[] Question: Does the paper describe potential risks incurred by study participants, whether such risks were disclosed to the subjects, and whether Institutional Review Board (IRB) approvals (or an equivalent approval/review based on the requirements of your country or institution) were obtained?
  \item[] Answer: \answerNA{} 
  \item[] Justification: The paper does not involve crowdsourcing nor research with human subjects.
  \item[] Guidelines:
    \begin{itemize}
      \item The answer NA means that the paper does not involve crowdsourcing nor research with human subjects.
      \item Depending on the country in which research is conducted, IRB approval (or equivalent) may be required for any human subjects research. If you obtained IRB approval, you should clearly state this in the paper.
      \item We recognize that the procedures for this may vary significantly between institutions and locations, and we expect authors to adhere to the NeurIPS Code of Ethics and the guidelines for their institution.
      \item For initial submissions, do not include any information that would break anonymity (if applicable), such as the institution conducting the review.
    \end{itemize}

  \item {\bf Declaration of LLM usage}
  \item[] Question: Does the paper describe the usage of LLMs if it is an important, original, or non-standard component of the core methods in this research? Note that if the LLM is used only for writing, editing, or formatting purposes and does not impact the core methodology, scientific rigorousness, or originality of the research, declaration is not required.
  \item[] Answer: \answerYes{} 
  \item[] Justification: In the appendix.
  \item[] Guidelines:
    \begin{itemize}
      \item The answer NA means that the core method development in this research does not involve LLMs as any important, original, or non-standard components.
      \item Please refer to our LLM policy (\url{https://neurips.cc/Conferences/2025/LLM}) for what should or should not be described.
    \end{itemize}

\end{enumerate}
\clearpage

\appendix
\section{Discussion and Derivation of Reg-GRPO}
\label{app_sec:derivation}
\subsection{Deriving the optimum of the KL-constrained reward maximization optimization}
We derive Eq.~\eqref{eq:pi} from Eq.~\eqref{eq:objective} following \cite{rafailov2023direct}.
We optimize $\pi_\theta$ with the following objective:
\begin{equation}
\label{eq_sup:objective}
    \pi_{\theta}^* = \underset{\pi_{\theta}}{\mbox{arg}\max}\  \mathbb{E}_{\boldsymbol{x},\boldsymbol{y} \sim \pi_{\theta}\left(\cdot| \boldsymbol{x}\right)} \mathcal{R}\left(\boldsymbol{x}, \boldsymbol{y} \right) - \lambda \; \mathbb{E}_{\boldsymbol{x}}\left[ \mathbb{D}_{\text{KL}}\left(\pi_{\theta}\left(\cdot | \boldsymbol{x} \right) || \pi_{{\theta_{\text{old}}}}\left(\cdot | \boldsymbol{x} \right) \right)\right],
\end{equation}
where $\mathcal{R}$ is the reward function, $\pi_{\theta_{\text{old}}}$ is the old policy model, and $\lambda~(\lambda \ge 0)$ denotes a hyperparameter.
We can obtain a closed-form solution to the above relative-entropy-regularized maximization problem.
\begin{equation}
\begin{split}
        \pi_\theta^* &=\underset{\pi_{\theta}}{\mbox{arg}\max}\ \  \mathbb{E}_{\boldsymbol{x}, \boldsymbol{y} \sim \pi_{\theta}\left(\cdot| \boldsymbol{x}\right)} \mathcal{R}\left(\boldsymbol{x}, \boldsymbol{y} \right) - \lambda \; \mathbb{E}_{\boldsymbol{x}}\left[ \mathbb{D}_{\text{KL}}\left(\pi_{\theta}\left(\cdot | \boldsymbol{x} \right) || \pi_{{\theta_{\text{old}}}}\left(\cdot | \boldsymbol{x} \right) \right)\right]\\
        &= \underset{\pi_{\theta}}{\mbox{arg}\max}\  \mathbb{E}_{\boldsymbol{x}, \boldsymbol{y} \sim \pi_\theta\left(\cdot | \boldsymbol{x} \right)}\left[\mathcal{R}\left(\boldsymbol{x}, \boldsymbol{y}\right) - \lambda \cdot \log \frac{\pi_\theta\left(\boldsymbol{y}|\boldsymbol{x} \right)}{\pi_{\theta_{\text{old}}}\left(\boldsymbol{y}|\boldsymbol{x} \right)} \right]\\
        &= \underset{\pi_\theta}{\mbox{arg}\min}\  \mathbb{E}_{\boldsymbol{x}, \boldsymbol{y} \sim \pi_\theta\left(\cdot | \boldsymbol{x} \right)}\left[ \log \frac{\pi_\theta\left(\boldsymbol{y}|\boldsymbol{x} \right)}{\pi_{\theta_{\text{old}}}\left(\boldsymbol{y}|\boldsymbol{x} \right) }-\frac{1}{\lambda} \cdot \mathcal{R} \left(\boldsymbol{x}, \boldsymbol{y}\right) \right]\\
        &= \underset{\pi_\theta}{\mbox{arg}\min}\  \mathbb{E}_{\boldsymbol{x}, \boldsymbol{y} \sim \pi_\theta\left(\cdot | \boldsymbol{x} \right)}\left[ \log \frac{\pi_\theta\left(\boldsymbol{y}|\boldsymbol{x} \right)}{\pi_{\theta_{\text{old}}}\left(\boldsymbol{y}|\boldsymbol{x} \right) \exp\left(\frac{1}{\lambda} \mathcal{R}\left(\boldsymbol{x}, \boldsymbol{y} \right) \right)} \right]\\
        &=\underset{\pi_\theta}{\mbox{arg}\min}\  \mathbb{E}_{\boldsymbol{x}, \boldsymbol{y} \sim \pi_\theta\left(\cdot | \boldsymbol{x} \right)}\left[ \log \frac{\pi_\theta\left(\boldsymbol{y}|\boldsymbol{x} \right)}{\frac{1}{Z\left(\boldsymbol{x}\right)}\pi_{\theta_{\text{old}}}\left(\boldsymbol{y}|\boldsymbol{x} \right) \exp\left(\frac{1}{\lambda} \mathcal{R}\left(\boldsymbol{x}, \boldsymbol{y} \right) \right)} - \log Z\left(\boldsymbol{x}\right) \right],
\end{split}
\end{equation}
where $Z\left(\boldsymbol{x}\right) = \sum_{\boldsymbol{y}}\pi_{\theta_{\text{old}}} \left(\boldsymbol{y}|\boldsymbol{x} \right) \exp \left(\frac{1}{\lambda}\mathcal{R}\left(\boldsymbol{x}, \boldsymbol{y} \right) \right)$ is a partition function.
Please note that the partition function is only dependent on $\boldsymbol{x}$ and the old policy $\pi_{\theta_{\text{old}}}$.

Now let $\bar{\pi}_\theta$ be defined as:
\begin{equation}
        \bar{\pi}_\theta\left(\boldsymbol{y} | \boldsymbol{x} \right) = \frac{1}{Z\left(\boldsymbol{x}\right)} \pi_{\theta_{\text{old}}}\left(\boldsymbol{y} | \boldsymbol{x} \right)\exp\left(\frac{1}{\lambda} \mathcal{R}\left(\boldsymbol{x}, \boldsymbol{y} \right) \right).
\end{equation}
It can be seen as a valid probability distribution as $\bar{\pi}_\theta \left(\boldsymbol{y}|\boldsymbol{x} \right) \ge 0$ for all $\boldsymbol{y}$ and $\sum_{\boldsymbol{y}}\bar{\pi}_\theta\left(\boldsymbol{y}|\boldsymbol{x} \right) = 1$.
Since $Z\left(\boldsymbol{x}\right)$ is not a function of $\boldsymbol{y}$, the above minimization problem can be formulated as
\begin{equation}
\begin{split}
        &\underset{\pi}{\mbox{arg}\min}\  \mathbb{E}_{\boldsymbol{x}, \boldsymbol{y} \sim \pi_\theta\left(\cdot | \boldsymbol{x} \right)}\left[ \log \frac{\pi_\theta\left(\boldsymbol{y}|\boldsymbol{x}\right)}{\bar{\pi}_\theta\left(\boldsymbol{y}|\boldsymbol{x}\right)} - \log Z\left(\boldsymbol{x}\right) \right]\\
        &=\underset{\pi}{\mbox{arg}\min}\  \mathbb{E}_{\boldsymbol{x}}\left[\mathbb{D}_{\text{KL}}\left(\pi_\theta\left(\boldsymbol{y}|\boldsymbol{x} \right) || \bar{\pi}_\theta\left(\boldsymbol{y}|\boldsymbol{x} \right) \right)  - \log Z\left(\boldsymbol{x} \right)\right]
\end{split}
\end{equation}
        
Since the partition function $Z\left(\boldsymbol{x}\right)$ is not dependent on $\pi$, the optimal $\pi^*$ is the policy that minimizes the first KL term.
Since the optimal KL-divergence is achieved if and only if two distributions are identical, we have optimal solution as:
\begin{equation}
\label{eq_sup:policy}
    {\pi}_{\theta}^*\left(\boldsymbol{y} | \boldsymbol{x} \right) = \frac{1}{Z\left(\boldsymbol{x}\right)} \pi_{\theta_{\text{old}}}\left(\boldsymbol{y} | \boldsymbol{x} \right)\exp\left(\frac{1}{\lambda} \mathcal{R}\left(\boldsymbol{x}, \boldsymbol{y} \right) \right),\quad \forall \boldsymbol{x}, \boldsymbol{y}.
\end{equation}

\subsection{Deriving the reward in terms of the optimal policy}
The reward can be reorganized under the optimal policy.
We can invert Eq.~\eqref{eq_sup:policy} as follows:
\begin{equation}
\label{eq_sup:reward}
\begin{split}
            \exp\left(\frac{1}{\lambda}\mathcal{R}\left(\boldsymbol{x}, \boldsymbol{y} \right)\right) &= Z\left(\boldsymbol{x}\right) \frac{\pi_{\theta}^*\left(\boldsymbol{y} | \boldsymbol{x} \right)}{\pi_{\theta_{\text{old}}}\left(\boldsymbol{y} | \boldsymbol{x} \right)},\\
        \frac{1}{\lambda}\mathcal{R}\left(\boldsymbol{x}, \boldsymbol{y} \right) &= \log Z\left(\boldsymbol{x} \right) + \log \left(\frac{\pi_{\theta}^*\left(\boldsymbol{y} | \boldsymbol{x} \right)}{\pi_{\theta_{\text{old}}}\left(\boldsymbol{y} | \boldsymbol{x} \right)} \right),\\
        \mathcal{R}\left(\boldsymbol{x}, \boldsymbol{y} \right) &= \lambda \cdot \left(\log Z\left(\boldsymbol{x} \right) + \log \left(\frac{\pi_{\theta}^*\left(\boldsymbol{y} | \boldsymbol{x} \right)}{\pi_{\theta_{\text{old}}}\left(\boldsymbol{y} | \boldsymbol{x} \right)} \right)\right), \quad \forall \boldsymbol{x}, \boldsymbol{y}.
\end{split}
\end{equation}

\subsection{Deriving the advantage in terms of the optimal policy.}

The advantage $\hat{A}^{(i)}$ is defined as
\begin{equation}
\label{eq_sup:advantage}
    \hat{A}^{(i)} =  \frac{\mathcal{R}\left(\boldsymbol{x}, \boldsymbol{y}^{(i)} \right) - \mu_r}{\sigma_r},
\end{equation}
where $\mu_r, \sigma_r$ denotes the average and standard deviation values of a set of rewards in the group, respectively.
We can rewrite the advantage in terms of the optimal policy in Eq.~\eqref{eq_sup:reward} as follows:
\begin{equation}
\begin{split}
\label{eq_sup:adv_approx}
\hat{A}^{(i)} &= \frac{   \rho^*\left(\boldsymbol{x},\boldsymbol{y}^{(i)} \right) + \log Z\left(\boldsymbol{x} \right) -  \left(\frac{1}{G}\sum_{j=1}^G  \rho^*\left(\boldsymbol{x},\boldsymbol{y}^{(j)} \right) + \log Z\left(\boldsymbol{x} \right) \right) }{\sigma_{\rho^*}},\\
&= \frac{   \rho^*\left(\boldsymbol{x},\boldsymbol{y}^{(i)} \right) - \mu_{\rho^*}}{\sigma_{\rho^*}},
\quad \rho^*\left(\boldsymbol{x}, \boldsymbol{y} \right) = \log \frac{\pi_{\theta}^*\left(\boldsymbol{y} | \boldsymbol{x} \right)}{\pi_{\theta_{\text{old}}}\left(\boldsymbol{y} | \boldsymbol{x} \right)},
\end{split}
\end{equation}
where $\mu_{\rho^*}, \sigma_{\rho^*}$ are mean and standard deviation of $\left\{ \rho^*\left(\boldsymbol{x}, \boldsymbol{y}^{(i)} \right) \right\}_{i=1}^G$, respectively.
Interestingly, we can see that $Z\left(\boldsymbol{x}\right)$ is removed.

\subsection{Reg-GRPO}
Based on Eq.~\eqref{eq_sup:adv_approx}, our Reg-GRPO~(Regressive GRPO) is to learn the model $\pi_\theta$ to directly predict the advantage as follows:
\begin{equation}
\begin{split}
& \mathcal{L}_{\text{Reg-GRPO}}\left(\theta \right) = \mathbb{E}_{\boldsymbol{x}, \left\{\boldsymbol{y}^{(i)} \right\}_{i=1}^G \sim \pi_{\theta_\text{old}}\left(\cdot | \boldsymbol{x}\right)} \left\{\left(\hat{A}^{(i)}- \hat{A}_\theta^{(i)}\right)^2 - \beta \; \mathbb{D}_{\text{KL}}\left[\pi_\theta || \pi_{\text{ref}} \right]\right\},\\
&\hat{A}^{(i)}_\theta = \frac{  \rho\left(\boldsymbol{x},\boldsymbol{y}^{(i)} \right) -  \mu_{\rho} }{\sigma_{\rho}}, \quad \rho\left(\boldsymbol{x}, \boldsymbol{y}\right) = \log \frac{\pi_\theta \left(\boldsymbol{y} | \boldsymbol{x} \right)}{\pi_{\theta_{\text{old}}} \left(\boldsymbol{y} | \boldsymbol{x} \right)},
\end{split}
\end{equation}
where $\mu_\rho, \sigma_\rho$ are the mean and standard deviation of $\left\{\rho\left(\boldsymbol{x}, \boldsymbol{y}^{(i)} \right) \right\}_{i=1}^G$, respectively.
For simplicity, we omit the KL divergence between the policy and the reference model.

\paragraph{Discussion.}
Motivated by Group-Relative Policy Optimization~(GRPO), the strength of Reg-GRPO lies in its regression-based approach to advantage estimation, unlike other methods that implicitly derive policy updates from preference probabilities.
By directly regressing the group-normalized target, Reg-GRPO leads to more precise updates, as the model is not just learning which response is better, but also how much better it is, according to the advantage.

In relation to Direct Preference Optimization~(DPO), Reg-GRPO offers a different perspective on leveraging preference data.
While DPO learns which response is preferred, Reg-GRPO attempts to capture a finer-grained signal about the degree of preference by directly regressing the advantages.
This could be particularly beneficial in scenarios where the difference in quality between preferred and dispreferred responses varies significantly.
Furthermore, the group-wise normalization inherent in GRPO, and carried into Reg-GRPO, can offer robustness when dealing with diverse and potentially inconsistently scaled preference data, which may require more careful handling in a pairwise DPO setup.

In contrast to REBEL~\cite{gao2024rebel}, one of the novel regression-based reinforcement-fine-tuning methods, which regresses the unnormalized pairwise reward gap differences between sampled outputs, our proposed Reg-GRPO framework directly learns to predict the group-based normalized advantage.
This shift from pairwise to group-level regression is a design choice that addresses the high variance typically observed in the outputs of video LLMs.
By normalizing log-probability ratios within each group, Reg-GRPO mitigates scale discrepancies across batches and enhances the optimization during training.
As a result, Reg-GRPO offers a more scalable and effective learning paradigm for fine-tuning language models using preference-based feedback.



\section{Reward functions}

To compute verifiable rewards, we follow existing works~\cite{feng2025video,chen2025exploring,li2025videochat} that fine-tune VideoLLMs with GRPO.

\noindent{\textbf{Format reward.}}
Following existing GRPO-based works~\cite{feng2025video,chen2025exploring,li2025videochat}, we employ a format reward to ensure that the model generates outputs in the desired format.
For example, the model is trained to output the thought process with \texttt{<think>...</think>} followed by the answer with \texttt{<answer>...</answer>}.
We use regular expressions to verify whether the outputs satisfy the specified format.
The format reward $R_{\text{format}}$ is applied to all tasks:
\begin{equation}
    R_\text{format}=
    \begin{cases}
        0, &\text{if output does not match the format}, \\
        1, &\text{if output matches the format}.
    \end{cases}
\end{equation}

\noindent{\textbf{Accuracy reward.}}
For tasks such as question answering, we employ an accuracy reward, which is formulated as:
\begin{equation}
    R_\text{acc}=
    \begin{cases}
        0, &\text{if}~\hat{a}\neq a \\
        1, &\text{if}~\hat{a}=a,
    \end{cases}
\end{equation}
where $a$ is the ground-truth answer and $\hat{a}$ is the model prediction, which is extracted from the regular expressions with \texttt{<answer>...</answer>}.

\noindent{\textbf{IoU reward for temporal perception.}}
We utilize an IoU reward to assess the model's ability to identify the temporal segment described by the input query and localize the target within the video.
The IoU reward is defined as:
\begin{equation}
    R_\text{IoU}=\frac{|\mathcal{P}\cap\mathcal{Q}|}{|\mathcal{P}\cup\mathcal{Q}|},
\end{equation}
where $\mathcal{P}$ and $\mathcal{Q}$ are the model prediction set and ground-truth set, respectively.
For the temporal grounding task, $\mathcal{P}$ and $\mathcal{Q}$ are defined as the timestamps of events within the video.

\section{Detailed Experimental Settings}
\label{app_sec:settings}

\subsection{Implementation Details}

\label{app_sec:impl_details}

We implement our code using the PyTorch library~\cite{paszke2019pytorch}.
We also adopt the Hugging Face Transformers library~\cite{wolf2019huggingface} and the TRL library~\cite{vonwerra2022trl} to post-train Video Large Language Models~(VideoLLMs).
For inference and rollout, we use vLLM~\cite{kwon2023efficient}.
For all the experiments, we fine-tune only a large language model while keeping the visual encoder frozen.
We use Qwen2.5-VL~\cite{bai2025qwen2} and Qwen2-VL~\cite{wang2024qwen2} as our base VideoLLMs.
We use NVIDIA A100 GPUs for 3B models and NVIDIA H200 GPUs for 7B models.
In addition, we use LLM-based tools for implementation and for correcting grammatical errors in the writing.

For the SEED-Bench-R1 dataset, we apply a KL-divergence regularizer between the model $\pi_{\theta}$ and the reference model $\pi_{\text{ref}}$ with coefficient 0.1, following prior GRPO works~\cite{shao2024deepseekmath,chen2025exploring}.
We use Qwen2.5-VL as the default base VideoLLM, and use Qwen2.5-VL-3B for all analyses.
We set the number of generations in the group as 8 for all the settings.
For DeepVideo-R1, we maintain a reward history using the most recent $W=100$ samples, and GRPO does not adopt safeguards based on our empirical study.
To train the model on the SEED-Bench-R1 dataset, we limit the maximum number of sampled frames per input video to 16 with a frame resolution of $252 \times 252$, and then append the frame indicating the current observation as an additional image input, following SEED-Bench-R1~\cite{chen2025exploring}.
To train the model using NExTGQA, we follow the experimental setups in VideoChat-R1~\cite{li2025videochat}.
\subsection{Evaluation Metrics}
\label{app_sec:metrics}

\noindent{\textbf{Accuracy.}} The accuracy metric measures the ratio of correct predictions that match the ground-truth answers for given questions, which is as follows:
\begin{equation}
    \text{Acc}=\frac{1}{N}\sum^{N}_{i=1}\mathds{1}\left(\hat{a}_i=a_i\right),
\end{equation}
where $N$ is the number of samples, $\hat{a}_i$ is the prediction, and $a_i$ is the ground-truth answer.

\noindent{\textbf{mIoU.}}
The mIoU (\ie mean Intersection over Union) metric calculates the average IoU over all samples, where IoU represents the similarity between the predicted and ground-truth timestamps, which can be formulated as:
\begin{equation}
    \text{mIoU}=\frac{1}{N}\sum^{N}_{i=1}\text{IoU}_i=\frac{1}{N}\sum^{N}_{i=1}\frac{|p_i\cap q_i|}{|p_i\cup q_i|}=\frac{1}{N}\sum^{N}_{i=1}\frac{|\left(s^p_i,e^p_i\right)\cap \left(s^q_i,e^q_i\right)|}{|\left(s^p_i,e^p_i\right)\cup \left(s^q_i,e^q_i\right)|},
\end{equation}
where $N$ is the number of samples, $p_i=\left(s^p_i,e^p_i\right)$ is the prediction, $q_i=\left(s^q_i,e^q_i\right)$ is the ground truth, and $s^p_i,e^p_i,s^q_i,e^q_i$ denote the start and end timestamps of the prediction and ground truth, respectively.

\noindent{\textbf{R@\textit{m}.}}
\cite{gao2017tall} proposed ``{$\text{R}@n,~\text{IoU}=m$}" metric for the temporal grounding tasks that measures the percentage of queries where at least one of the top-$n$ predictions has an IoU higher than $m$ with the ground-truth.
Following~\cite{li2025videochat}, we adopt $\text{R}@m$ as a top-1 variant of ``$\text{R}@n,~\text{IoU}=m$'', defined as:
\begin{equation}
\begin{split}
    \text{``R}@n,~\text{IoU}=m\text{''}&=\frac{1}{N}\sum^{N}_{i=1}\mathds{1}\left(\text{IoU}^j_i\ge m,~\exists j\in\{1,2,\dots,n\}\right), \\
    \text{R}@m=\text{``R}@1,~\text{IoU}=m\text{''}&=\frac{1}{N}\sum^{N}_{i=1}\mathds{1}\left(\text{IoU}_i\ge m\right),\quad \text{where}~\text{IoU}_i=\text{IoU}^1_i,
\end{split}
\end{equation}
where $N$ is the number of samples, $m$ is the IoU threshold, and $\text{IoU}^j_i$ is the IoU between the $j$-th prediction (ranked among the top-$n$ predictions) and the ground truth.

\section{RL Baselines}
\label{app_sec:baselines}
In this section, we describe the baseline methods used to compare against Reg-GRPO in Table~\ref{tab:analysis_on_RFT} of the main paper.

\noindent\textbf{DPO~\cite{rafailov2023direct}} aligns model outputs with human preferences using pairwise comparisons.
For DPO, we sample model outputs from a fixed reference model.

\noindent\textbf{Online DPO~\cite{rafailov2023direct}} also adopts direct preference optimization to learn the model.
Compared to standard DPO, it samples outputs from the old policy model, which evolves during training, following GRPO~\cite{shao2024deepseekmath}.

\noindent\textbf{REINFORCE~\cite{williams1992simple,kreutzer2017bandit,nguyen2017reinforcement}} is a classic policy-gradient algorithm that updates the model using the reward-weighted log-likelihood of the outputs.
Generally, it also samples the outputs from the old policy model~$\pi_{\theta_{\text{old}}}$.

\noindent\textbf{REINFORCE Leave-One-Out~(RLOO)~\cite{kool2019buy}} is designed to reduce the variance of gradient estimates in REINFORCE.
Instead of using the reward directly as a weight, it uses a Monte Carlo estimate to compute a baseline and subtracts it from the reward when forming the weight.
Same as REINFORCE, it samples the outputs from the old policy model~$\pi_{\theta_{\text{old}}}$.

\noindent\textbf{REBEL~\cite{gao2024rebel}} directly regresses the pairwise reward gap, which motivates us to apply the regression-based fine-tuning methods.
Different from our work that directly predicts the group-normalized advantage, it regresses the unnormalized pairwise reward gap.

\noindent\textbf{Reward-Regression~(Eq.~(5))} is our baseline that directly regresses the reward by approximating $Z\left(\boldsymbol{x}\right)$ with Monte-Carlo sampling.
Since $Z\left(\boldsymbol{x}\right)$ is not accurate and relying solely on the reward introduces high variance, it performs worse than our Reg-GRPO.

\section{Datasets}
\label{app_sec:datasets}

\noindent{\textbf{SEED-Bench-R1}}~\cite{chen2025exploring} is a dataset designed to evaluate the effectiveness of post-training methods in the context of video understanding capabilities of MLLMs.
Specifically, the dataset incorporates Epic-Kitchens~\cite{damen2022rescaling} and Ego4D~\cite{grauman2022ego4d} as videos and EgoPlan-Bench~\cite{chen2023egoplan} and EgoPlan-Bench2~\cite{qiu2024egoplan} as benchmark sources to construct a hierarchical validation structure, enabling evaluation across diverse real-world scenarios.

\noindent{\textbf{LongVideoBench}}~\cite{wu2024longvideobench} contains 3,763 videos and 6,678 QA pairs, where videos are diverse in domain~(\eg Life, Movie), task~(\eg scene-referred event, object before/after text), and duration.
In particular, the video durations are divided into four progressive groups, \textit{(8s, 15s], (15s, 60s], (180s, 600s], (900s, 3600s]}, with an overall average of 100s, facilitating the assessment of the model’s understanding of long-context interleaved multimodal inputs.

\noindent{\textbf{VSI-Bench}}~\cite{yang2024thinking} is a dataset proposed to evaluate the visual-spatial intelligence capabilities of MLLMs, comprises over 5,000 question-answer pairs and 288 real videos.
Specifically, eight tasks~(object count, relative distance, relative direction, route plan, object size, room size, absolute distance, appearance order) of three types~(configurational, measurement estimation, spatiotemporal) are defined within the dataset.

\noindent{\textbf{Video-MMMU}}~\cite{hu2025video} consists of 300 expert-level videos and 900 questions, targeting the evaluation of the knowledge acquisition capabilities in MLLMs.
Inspired by the human process of acquiring knowledge to solve challenging problems, the questions in the dataset are human-annotated across six disciplines~(Art, Business, Science, Medicine, Humanities, Engineering) and aligned with three stages: Perception, Comprehension, and Adaptation.

\noindent{\textbf{MMVU}}~\cite{zhao2025mmvu} comprises 3,000 expert-annotated question-answer pairs and 1,529 specialized-domain videos covering 27 subjects across 4 key disciplines~(Science, Healthcare, Humanities \& Social sciences, Engineering), and aims to evaluate the expert-level, knowledge-intensive video understanding abilities of MLLMs.
Following~\cite{feng2025video}, we report the performance on multiple-choice QA.

\noindent{\textbf{MVBench}}~\cite{li2024mvbench} serves as a benchmark to assess temporal comprehension capabilities of MLLMs, and consists of 20 challenging video understanding tasks that require reasoning beyond a single frame.
In particular, the dataset is built upon videos sourced from various benchmarks, enabling the evaluation of MLLMs' general ability for open-world temporal understanding.
Since each task contains 200 question-answer pairs, we conduct evaluation on a total of 4,000 question-answer pairs.

\noindent{\textbf{TempCompass}}~\cite{liu2024tempcompass} is designed for evaluating the temporal perception ability of MLLM, based on 5 basic temporal aspects~(Action, Speed, Direction, Attribute change, Event order) and 10 fine-grained sub-aspects~(\eg relative speed, camera direction, combined change).
We report results on overall performance, including all four tasks~(Multi-choice QA, Yes/No QA, Caption matching, Caption generation), for comparison with prior work~\cite{feng2025video}.

\noindent{\textbf{Video-MME}}~\cite{fu2024video} is a dataset for evaluating the general video understanding capabilities of MLLMs, consisting of 900 videos and 2,700 question-answer pairs, where the videos are constructed with variation in both type and temporal duration.
Specifically, the dataset covers 6 key domains and 30 sub-class video types, and each video is categorized as short~(< 2 mins), medium~(4-15 minutes), or long~(30-60 minutes) depending on its duration length.
We report the average performance across all temporal duration splits, without using subtitles.

\noindent{\textbf{NExTGQA}}~\cite{xiao2024can} is a temporal grounding QA dataset consisting of 5,417 videos, 43,043 QA pairs, and 10,531 timestamp labels.
As temporal segment annotations are available only in the validation and test splits, we use the validation and test splits for the model training and model evaluation, respectively.

\section{Broader Impacts and Limitations}

\subsection{Broader Impacts}
\label{app_sec:impact}

We propose a video large language model named DeepVideo-R1, trained with Regressive GRPO~(Reg-GRPO) and difficulty-aware data augmentation.
DeepVideo-R1 is broadly applicable to complex video reasoning tasks.
We believe that DeepVideo-R1 itself does not introduce new negative impacts.
However, as the model is based on pretrained language and vision models, it may generate biased outputs related to race, religion, culture, and gender, which could lead to misuse.
In addition, training VideoLLMs may result in CO$_2$ emissions, which contribute to global warming.
\subsection{Limitations}
\label{app_sec:limitations}

Our DeepVideo-R1 is built upon a large-scale pretrained video large language model and fine-tuned on video reasoning datasets to leverathe the rich world knowledge embedded in the pretrained models.
However, it remains unclear whether there is any overlap between pertaining content and downstream evaluation benchmarks.
This uncertainty may introduce a risk of implicit data leakage.
Furthermore, since DeepVideo-R1 is based on VideoLLM, it has following limitations: high computational and memory requirements.
As DeepVideo-R1 is fine-tuned on top of such models, it may inevitably inherit these challenges.

\end{document}